# Electrically-driven phase transition actuators to power soft robot designs

D. Fonseca,[1] P. Neto[1]*

[1]*University of Coimbra, CEMMPRE, Department of Mechanical Engineering, Coimbra, Portugal*

**Corresponding author. Email: pedro.neto@dem.uc.pt*

Supplementary materials: https://softrobotic.github.io/untapped_potential/

**Abstract:** In the quest for electrically-driven soft actuators, the focus has shifted away from liquid-gas phase transition, commonly associated with reduced strain rates and actuation delays, in favour of electrostatic and other electrothermal actuation methods. This prevented the technology from capitalizing on its unique characteristics, particularly: low voltage operation, controllability, scalability, and ease of integration into robots. Here, we introduce a liquid-gas phase transition electric soft actuator that uses water as the working fluid and is powered by a coil-type flexible heating element. It achieves strain rates of over 16%/s and pressurization rates of 100 kPa/s. Blocked forces exceeding 50 N were achieved while operating at voltages up to 24 V. We propose a method for selecting working fluids which allows for application-specific optimization, together with a nonlinear control approach that reduces both parasitic vibrations and control lag. We demonstrate the integration of this technology in soft robotic systems, including a cable-driven biomimetic hand and a quadruped robot powered by liquid-gas phase transition.

## INTRODUCTION

Nature is a rich source of inspiration for soft roboticists [1,2], who often learn from observing various forms of animal and plant life thereby creating innovative robot designs [3–7]. This bioinspired approach steers research towards tackling complex challenges such as locomotion and manipulation in unstructured environments. Soft robots are becoming ever more complex systems, dependent on multiple core technologies, including: materials, fabrication processes [8–10], sensors [11,12], electric conductors [13,14], and actuators [1,15]. Various energy sources have been used to power soft actuators, namely: pressurized fluids [16–19], chemical energy [6,20–22], heat [7,23–25], magnetic or electric fields [26–29] and electric current [30,31].

Electrically-driven soft actuators are of particular importance due to the practicalities of storing, transmitting, and controlling electric energy. The use of electronic circuits to control soft robots enables logic and communication capabilities beyond those of on-board microfluidic logic systems [32–35]. Electric soft actuation has been achieved through various methods. Electrostatic actuators, such as Dielectric Elastomer Actuators (DEAs) and Hydraulically Amplified Self-Healing Electrostatic Actuators (HASELs) [36–38], utilize Maxwell stresses to deform their structure. Piezoelectric actuators rely on the reverse piezoelectric effect to generate work [39,40]. Electrochemical actuators rely on pressure gradients resultant from electrolysis [41], intercalation processes [42–44] or redox reactions [45–47]. Electrothermal actuators use heat as an intermediary form of energy that is then converted into work through thermal expansion or phase transitions. Given the unique features and characteristics of each actuation technology, we believe it is key to develop them all to a stage at which their full potential can be thoroughly evaluated.

Let us consider electrothermal actuators. Compared to their electrostatic counterparts, they tend to operate at lower voltages (Fig. 1a), and are usually resistive electric loads [48]. This combination not only reduces the risk of electromagnetic interference but also allows for safer and simpler driver circuits. Electrothermal actuators can also achieve some of the highest work densities [1,16], making them particularly attractive for applications requiring actuation in compact designs. However, these benefits come with well-known trade-offs. Electrothermal actuation typically suffers from low strain rates (Fig. 1b), preventing them from being deployed in high-speed applications. Furthermore, technologies based on fiber-like thin structures, common in Shape Memory Alloys (SMAs), Twisted-and-Coiled Actuators (TCAs) [49], and Liquid Crystal Elastomers (LCEs) [50], can be troublesome to scale up due to limitations regarding physical integrity and thermal stress. Safety concerns arising from high operating temperatures must also be considered when implementing electrothermal soft actuators.

In this work, we focus on the particular case of liquid-gas phase transition actuators, where a liquid working fluid is heated and vaporized, creating pressure that deforms soft structures [51]. Since the working fluid provides uniform pressure distribution, liquid-gas actuators are potentially less susceptible to thermal stress and offer higher design flexibility when compared to other phase transition actuators such as LCEs, where actuation is dependent on the alignment of liquid crystal molecules [52]. This design flexibility is demonstrated in Figs. 1c and 1d, where actuators with different actuation modes can share the same internal components. The decoupling between pressure-generating components, heating element and working fluid, and external soft structures also allows actuators to be optimized to achieve highly specific actuation modes. This design ensures they are robust enough to operate in direct exposure to the outside environment, eliminating the need for additional protection provided by the robot's main structure.

However, liquid-gas phase transition actuators exhibit significant limitations. They feature low strain rates, often with multi-second delays, even when compared to other electrothermal technologies (Fig 1b). Contractile strain rates sit currently in the range of 1-2%/s [53–55]. They also face rapid degradation of performance due to the depletion of the working fluid by diffusion through the actuator's structure [56,57]. Surprisingly, our experimental findings also revealed that these actuators are subject to mechanical vibrations, an issue that has been seemingly overlooked in previous literature. We demonstrate that these instabilities stem from operating in thermodynamically unsaturated boiling regimes [58,59] and can be significantly mitigated through the implementation of appropriate control strategies. We also demonstrate how these same control strategies can effectively reduce actuation delays.

Apart from addressing existing limitations, we set out to explore the untapped potential of liquid-gas phase transition. We start by introducing an innovative actuator design (Figs. 1d and 1e), and propose a systematic approach to selecting the working fluid that enables application-specific temperature and pressure optimization. The selection of working fluids opens an opportunity to fine-tune operating temperatures to meet specific application requirements. Although no comprehensive methods for working fluid selection have been previously proposed in the context of soft actuators, foundational experimental research has already been outlined [60].

To further enhance performance, a nonlinear control strategy is proposed, capable of effectively mitigating parasitic vibrations and reducing control lag. We then investigate pressurization limitations, including the limit imposed by Critical Heat Flux (CHF), Fig. 1f, demonstrating a pressurization rate of 100 kPa/s. These performance figures suggest the feasibility of a class of high-performance liquid-gas phase transition actuators.

Finally, we demonstrate how the proposed actuator design can be integrated into compliant mechanical systems by developing a cable-driven biomimetic hand, a soft robotic gripper, and Bixo, a quadruped robot powered by liquid-gas phase transition actuators, Figs. 1g to 1j. The aim of this work is to provide a good understanding of the untapped potential and limitations of this underexplored actuation technology, which we believe will play a significant role in the development of future untethered soft robots [16].

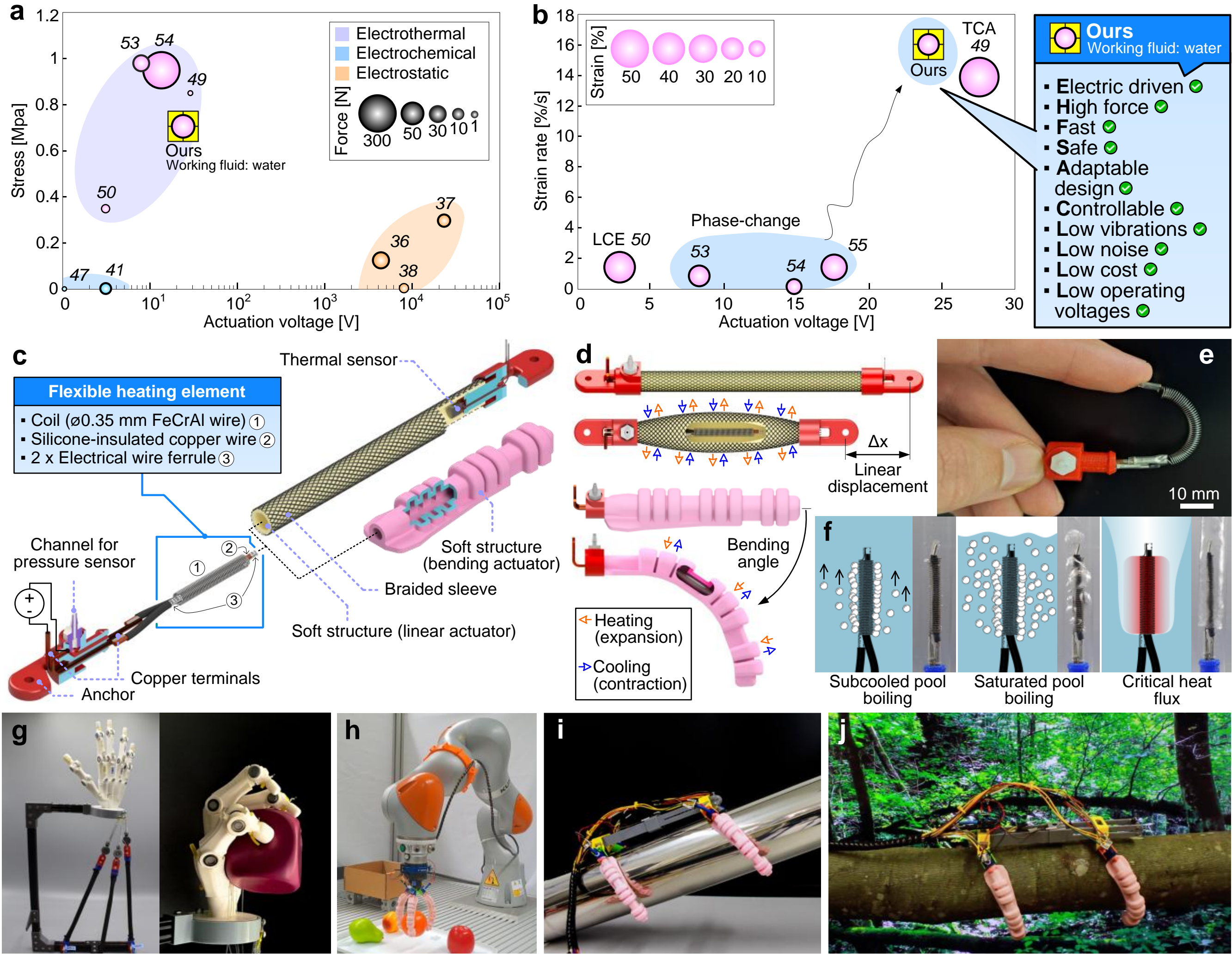

**Fig. 1 | Liquid-gas phase transition actuator: comparative performance, design, working principle and robot prototypes. a** Literature survey of electric linear soft actuators (stress and force with respect to actuation voltage). **b** Literature survey of electrothermal linear contractile actuator performance (strain and strain rate with respect to actuation voltage). **c** Design and exploded view of the linear and bending actuators. **d** Actuation principle based on pressure generated through liquid-gas phase transition. **e** Actuator core with flexible heating element. **f** Operating regimes: subcooled pool boiling, saturated pool boiling and critical heat flux. **g** Biomimetic hand powered by linear phase transition soft actuators. **h** Soft robotic gripper powered by bending phase transition soft actuators. **i** Bixo robot powered by four phase transition actuators, crawling on an inclined metallic tube. **j** Bixo robot crawling on a horizontal tree trunk.

## RESULTS

### Design guidelines

To unlock a class of high-performance liquid-gas phase transition actuators, a critical review of previous design concepts is required. While maintaining the underlying principle of creating pressure inside a soft structure by boiling a working fluid, different methods have been proposed, posing multiple questions whose answers shape the design. Working fluids can be held in bulk inside a pressure chamber [55] or suspended in a silicone matrix [56]. Heating elements may or may not be designed to have direct contact with the working fluid [7,27]. The electric energy required to power the heating element may be wired through internal conductors or transmitted wirelessly [53]. Useful work can be generated using various soft structures, including McKibbens, bellows, pneu-nets and pouches.

Our design concept (Fig. 1c) reflects a prioritization of high heat flux capacity and design modularity. The working fluid is stored in bulk to avoid the thermal mass of an inert matrix. The heating element is in direct contact with the fluid to reduce the thermal resistance between the two. Additionally, electric power is directly wired to the heating element using internal conductors, which simplifies the overall actuation system and avoids the losses associated with wireless power transmission. In the proposed linear actuator variant, we included the option for two redundant state variable sensors, both an internal temperature and an internal pressure sensor. However, as will be shown and discussed shortly, we later concluded that pressure-based control is preferable, so we refrained from including temperature sensors in the bending actuators, which allowed for an even simpler design. A pressure tap, used to connect the pressure sensor, also allows for the working fluid to be replaced or replenished, if necessary, effectively removing the main life-limiting factor of previous designs.

Some of the previous design decisions raise concerns regarding reliability. The use of a working fluid in bulk implies that actuators will fail in the event of a single leak, while direct contact between the heating element and the fluid increases the risk of oxidation or chemical incompatibility. The routing of electric conductors into the pressurized soft structure also introduces pathways and joint interfaces that are prone to leaks. These concerns were addressed by following a modular design approach that enabled rapid iteration at the component level, as well as using monolithic elastomer structures for all the soft structures. The linear actuator variant is based on the well-established McKibben design, due to its simplicity and robustness, while the bending actuator variant is based on a fast Pneu-Net (fPN) design (Figs. 1c and 1d), which is one of the highest performing geometries for pneumatic soft bending actuators [61]. Details on the design and fabrication of all actuator components, including the heating element and soft structure, are available in Supplementary Notes 2 and 3.

**Actuation principle**

The boiling of a bulk mass of fluid confined within a hermetically sealed soft actuator can be approximated as a form of pool boiling. When the heating element's surface temperature rises above the local saturation temperature of the fluid, vapor bubbles nucleate and increase in size until buoyant forces cause them to detach. These gas bubbles are initially not in thermodynamic equilibrium with the adjacent liquid phase, which results in a transfer of heat between the two phases, liquid and gas. While the temperature of the liquid phase is kept lower than its saturation temperature, vapor bubbles tend to collapse before they reach the liquid's free surface, a regime known as subcooled pool boiling (Fig. 1f and Supplementary Movie 1). When the liquid temperature gets sufficiently close to saturation, vapor bubbles will rise all the way to the surface, entering a regime known as saturated pool boiling. The effect of bubble collapse under subcooled pool boiling is similar to the phenomena of cavitation, and results in pressure shock waves that compromise the integrity of the actuator. These bubble collapses are also the main source of instability causing mechanical vibrations in the soft actuator. To avoid premature failure and reduce the magnitude of actuator vibrations, we propose a control strategy which aims to maintain actuators in thermodynamic saturation during the entire actuation cycle. Vibrations, or any other form of actuation instability, significantly impact the implementation of actuators in delicate mechanical systems, including soft robots. Additionally, vibrations are a primary source of acoustic noise, which can be a critical factor for operations in noise sensitive environments.

An effective controller should also make full use of the actuator's heat flux capacity, providing as much peak power as the actuator can safely handle. The thermodynamic limit for this heat transfer rate is reached when an excess of vapor bubbles coalesce, creating a gaseous layer that separates the heating element's surface from the liquid phase. At this point, the thermal resistance increases significantly as heat is now being transferred through a solid-gas interface and, as a result, the surface temperature of the heating element will increase accordingly, compromising the integrity of the system. The heat flux at this point is known as the Critical Heat Flux (CHF) and sets a practical limit to the pressurization rate of liquid-gas phase transition soft actuators [62].

**Working Fluid Selection**

The selection of the working fluid is an important opportunity to fine-tune actuators to meet specific application requirements. We introduce a four-step selection method based on thermodynamic and safety data. Step one involves selecting fluids based on the desired actuator operating pressure and temperature ranges. We started by excluding fluids that are not liquid at ambient temperature and pressure (20ºC, 101.33 kPa). We then selected an upper temperature limit of 138ºC, which is the softening temperature of the actuator's polyurethane components. A maximum relative pressure limit of 130 kPa was also defined, based on a safety factor of 2 applied to early destructive test data. Fluids outside these ranges are excluded (Fig. 2a), specifically R-11 and toluene.

Step two involves an evaluation of fluid-material compatibility. Fluorine-based fluids, including the Novec series of engineering fluids (3M, USA), as well as ethanol, are known to swell and diffuse through silicones and natural rubber [63–66]. Water has been demonstrated to have lower diffusion rates through silicone [64], suggesting a longer life expectancy for the actuators. Accordingly, in this step we excluded all fluorinated fluids and ethanol.

Step three assesses fluid safety, toxicity, and sustainability. Fluids with low flashpoints, such as acetone or ethanol pose fire hazards, and some have significant acute toxicity. Available information on toxicity parameters and symptoms of exposure is compiled in Supplementary Table 4 and 5. The simple, robust design of these actuators makes them suitable for unprotected use in various environments, making their sustainability heavily dependent on selecting fluids with minimal environmental impact. A compilation of the ecotoxicological data available is shown in Supplementary Table 6. This step identified water as the safer alternative.

Step four benchmarks' fluids using a custom metric we named the boundary work coefficient ($\omega_b$). This coefficient is the ratio of work done by the expanding fluid to the heat provided, assuming an idealized isobaric expansion. It can be calculated from the fluid's pressure-temperature saturation curve, critical temperature, enthalpy of vaporization, and molar weight. Results are shown in Figs. 2b and 2c. A comprehensive step-by-step example of the calculation procedure is detailed in the Supplementary Note 1. While $\omega_b$ does not represent actual actuation efficiency, it is useful for comparing fluids. Data suggest a negative correlation between enthalpy of vaporization and $\omega_b$, so we recommend giving preference to fluids with lower enthalpy of vaporization. While recent experimental observations have shown no evident correlation between the enthalpy of vaporization and maximum actuation strain [60], selecting fluids with a higher enthalpy of vaporization could allow for higher CHF values [67] and, potentially, increase the maximum pressurization rate at which a given actuator design may operate. However, as we will demonstrate, the current limiting factor for actuation cycle speed is the rate of condensation during the cool-down period, which is not directly related to the CHF. Moreover, for our actuators, we prioritized safety, selecting low-toxicity, non-flammable fluids without hazardous byproducts, low mass diffusivity through silicone, and ample thermodynamic data availability, leading us to choose water. The higher enthalpy of vaporization of water suggests our performance figures are conservative baselines for future high-performance actuators.

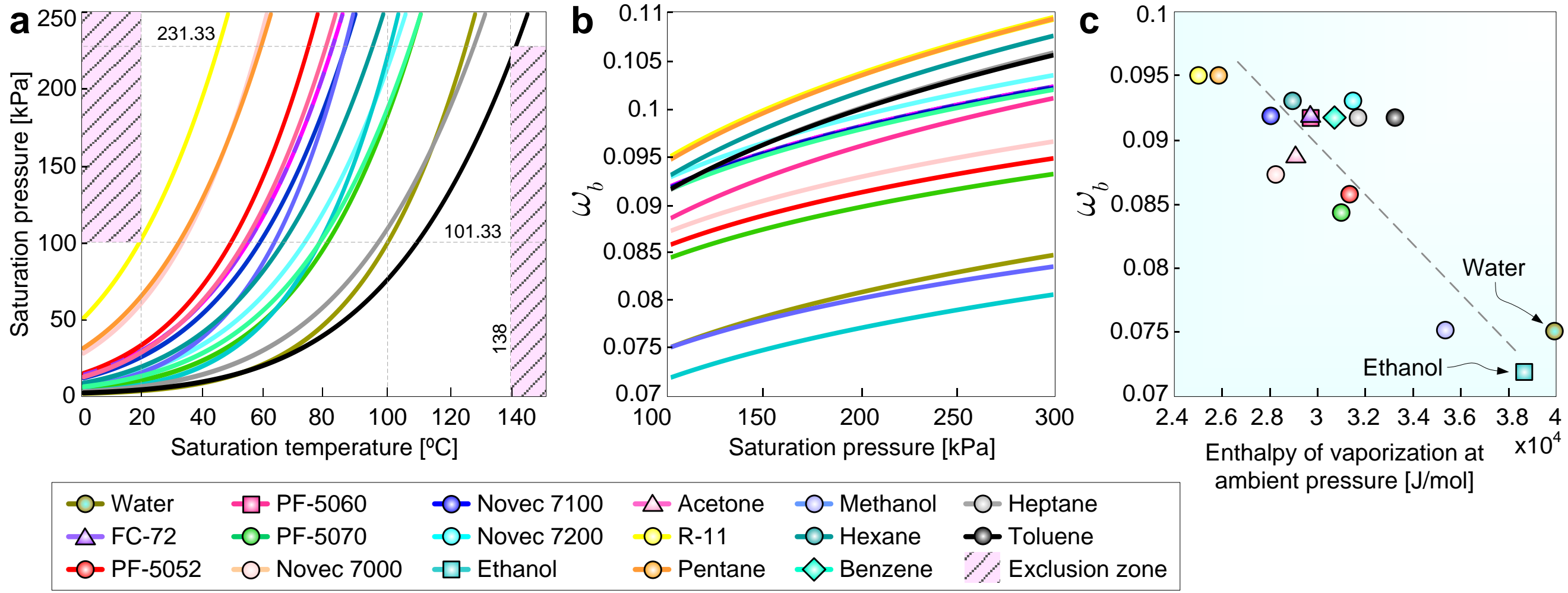

**Fig. 2. | Fluid data for the five-step selection method. a** Saturation pressure relative to the saturation temperatures. **b** Boundary work ratio relative to the saturation pressure. The boundary work ratio is the ratio of the boundary work done by the expanding fluid to the heat provided. **c** Boundary work ratio relative to the enthalpy of vaporization at ambient pressure. The dashed line illustrates the underlying trend observed across the different fluids.

### Influence of elastomer hardness

The level of decoupling between the different actuator components, characteristic of our modular design, creates an opportunity to develop actuator soft structures with a greater focus on the application requirements, while being subjected to fewer constraints internal to the actuator. In particular, the separation between the work generating component (working fluid) and the soft structure itself, allows us to approach the selection of elastomer materials with a focus on increasing performance figures such as strain rates, actuation force, or stability under external loads.

To verify the influence of the elastomer material on actuation performance we fabricated four geometrically identical linear actuators, each of them using a different elastomer for their soft structure. The chosen elastomers consisted of a variety of platinum-cured silicones (Ecoflex 00-30, Ecoflex 00-50, and Smooth-Sil 940 from Smooth-On, USA) alongside natural rubber. This material selection spans a range of hardness levels extending from Shore 00-30 to Shore 00-80. The actuators were then subjected to isometric and isotonic testing to evaluate differences in performance. For the isometric testing setup, each actuator was driven by a Proportional Integral (PI) controller which was commanded to pressurize it at a rate of 1 kPa/s (Fig. 3a). A linear pressure-force response was observed across all the actuators, with force slopes monotonically increasing with decreasing material hardness (Fig. 3b). This effect can be leveraged to adjust the actuator's operating pressures according to the application requirements without the need for geometric design adjustments. It also highlights the importance of selecting elastomer material in conjunction with the working fluid to ensure that both the operating pressures and temperatures meet the application requirements. The maximum blocked force achieved during the test was 52 N, achieved by the Ecoflex 00-30 actuator at 308 kPa (Fig. 3b).

We then evaluated isotonic performance by performing five cycles of pressurization/depressurization, also at a rate of 1 kPa/s while acting against a constant load of 10 N (Fig. 3c). Comparing the force gains allowed by softer materials, observable in Fig. 3b, to the displacement gains allowed by the same materials in Fig. 3c leads us to conclude that the use of softer elastomers increases displacement more than they increase force. As a result, we selected the softest material (Ecoflex 00-30) to fabricate all subsequent linear actuators. Additionally, no significant hysteresis was observed in the isotonic test, consistent with the behavior of traditional pneumatic McKibben actuators [68,69]. Under no load conditions, a linear actuator made of Ecoflex 00-30 achieved a contractile strain rate of 16.6%/s. The actuator proceeded to reach a maximum unloaded contractile strain of 27%, which is consistent with the performance of pneumatic McKibben soft actuators [70].

We then used the actuator made of Ecoflex 00-30 to obtain more comprehensive isotonic data by subjecting it to various loads while measuring their maximum displacement (Fig. 3d). Tests at higher loads were extended to 200 kPa, allowing us to achieve a realistic assessment of the actual work output capability, unconstrained by our empirically defined safety limit of 130 kPa. During these tests, the maximum work output was 0.56 J, corresponding to a specific energy of approximately 30 J/kg and an energy efficiency of 0.11%. Regardless of power settings, the actuator reached peak efficiency when actuating against an 18 N load, as detailed in Supplementary Note 7. Finally, the selection of the material for the soft structure of the bending actuator needed to ensure sufficient stiffness to avoid unwanted bending under the influence of gravity. We settled for Smooth-Sil 940 due to its higher hardness value (Shore 00-80). A flat membrane was also included in the design to increase structural stiffness under lateral loads. The bending coefficient, a metric proposed to compare different bending actuator designs, and which consists of the ratio of curvature over input pressure [71], remained relatively stable with an average value of $0.5\times10^{-3}$ m/N at pressures up to 50 kPa (Fig. 3e). At 50 kPa the actuator reaches its maximum design bending angle of 90 degrees.

**Performance limits**

We proceeded to evaluate the limits of the proposed design by subjecting a linear soft actuator to an isometric shock test. Starting from a preheated state, the actuator was supplied with a short-term electric impulse corresponding to 110 W of electric power. During this test, a pressurization rate of 100 kPa/s was achieved, (Fig. 3f). The actuator experienced an overpressure failure at 210 kPa without any signs of burned elements, suggesting that CHF conditions may not have been achieved even at these power settings. At 110 W, the heat flux on the heating element's surface is approximately 213 kW/$m^2$. We then tested the same heating element inside a transparent glass tube, under conditions similar to those expected inside the actuator, to visually confirm the onset of CHF conditions. In this case, CHF was achieved only at approximately 388 kW/$m^2$ (Fig. 3h). Knowing the maximum pressurization rate to be at least 100 kPa/s, we then pressurized an identical prototype to the same pressure of 210 kPa and recorded its internal pressure during the cooling down phase (Fig. 3g). The highest magnitude of depressurization rate, 35.8 kPa/s, was measured at 210 kPa, reducing to 7.1 kPa/s at 130 kPa and 0.5 kPa/s at 20 kPa. The results demonstrate the achievable pressurization performance and highlight a limiting factor, the rate of heat dissipation.

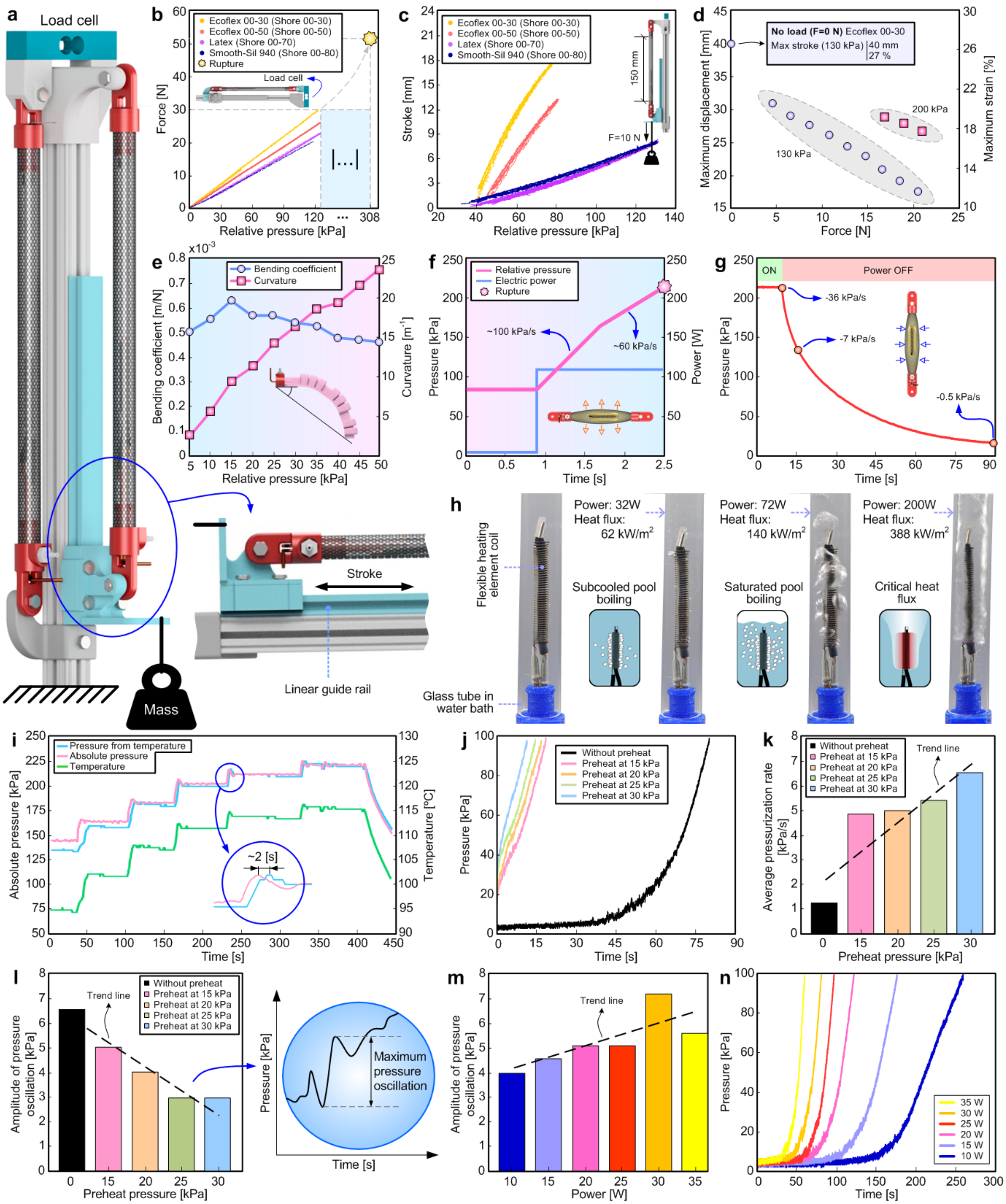

**Fig. 3. | Experimental tests and actuator characterization.** **a** Setup used for isometric and isotonic tests. **b** Blocked force of linear actuators made using different elastomers, tested under isometric conditions without prestrain. **c** Strain of linear actuators under a constant load of 10 N. **d** Maximum

strain of a linear actuator made from Ecoflex 00-30 under multiple loads while operating at two different pressures. **e** Curvature and bending coefficient of a bending actuator. The bending coefficient was calculated in accordance with the method originally proposed by Gorissen et al. [71]. **f** Electric power, internal pressure and internal pressure rate of a linear actuator during an impulse shock, tested under isometric conditions without prestrain. **g** Internal pressure during the cooling down of a linear actuator, tested under isometric conditions without prestrain. **h** Demonstration of boiling regimes occurring at different heat fluxes. Tested using an actuator heating element and a water-filled transparent glass tube made to be geometrically identical to an actuator's silicone soft structure. **i** Saturation pressure calculated from internal temperature measurements compared against actual internal pressure. **j** Pressure evolution at a 25 W constant power setting, with and without preheating. Test under isometric conditions without prestrain. **k** Average pressurization rate at 25 W, with and without preheating. Test under isometric conditions without prestrain. **l** Amplitude of pressure oscillations at 25 W, with and without preheating, and tested under isometric conditions without prestrain. The inset shows the definition of pressure oscillation. **m** The amplitude of a linear actuator's pressure oscillations at different power settings under isometric conditions without prestrain. **n** Internal pressure of a linear actuator, at different power settings, under isometric conditions without prestrain. Unless otherwise stated, all tests were conducted with linear actuators made from Ecoflex 00-30.

### Control strategy

Previous liquid-gas phase transition actuators have been controlled using PI and ON-OFF controllers, with feedback based on pressure, strain, force or temperature [51,53–55,72]. Determining an appropriate feedback signal for a class of high-power actuators is a key factor. In a McKibben actuator, force is a function of both strain and pressure. Therefore, a single feedback signal of force or strain would not allow for the effective detection of overpressure events.

Temperature and pressure are both thermodynamic state variables, so one can be derived from the other assuming steady state and saturated fluid conditions. We evaluated the feasibility of temperature-based control by powering up a linear actuator under isometric conditions while capturing internal temperatures and pressures using the actuator's built-in sensors. We then calculated the fluid saturation pressure from the temperature measurements. An offset of 13 °C was used during this calculation to account for internal temperature gradients. This offset adjustment allows for reasonable pressure tracking using temperature measurements, but temperature signals lagged behind pressure feedback by approximately 2 s due to the thermal masses present (Fig. 3i). Having demonstrated the potential for pressurization rates of 100 kPa/s, we consider feedback signal lags over 100 ms to be incompatible with high power operation. The conclusion is that pressure feedback is, in fact, the preferred option.

We also evaluated whether the onset of CHF could be detected by monitoring the electrical resistance of the actuator. Our hypothesis was that a significant temperature increase in the coil's surface would result in an abrupt variation in the coil's electrical resistance value, which would be detected by the controller and used to limit electrical supply during high-power short-term impulses. We achieved mixed results. Data recovered during the CHF experiment in Fig. 3h, where the coil was statically assembled inside a rigid test chamber, does show an abrupt change in electrical resistance occurring at the precise instant when we can visually confirm the onset of CHF. However, when analysing similar data obtained from real actuators, we could not reliably observe this phenomenon due to an excess of noise in the data. Although we could not determine the origin of this noise, it may be related to thermal strain of the crimp connections used to connect the coil to the internal electrical connections (Supplementary Note 2).

An effective control strategy must also ensure that the actuators are kept in thermodynamically saturated conditions. We achieved this by setting a minimum standby pressure that is higher than the atmospheric pressure. Four different standby pressure values were tested, ranging from 0 kPa to 30 kPa, by pressurizing a linear actuator under a constant power of 25 W for 60 s. The results showed that preheating effectively reduces control lag (Fig. 3j), increases pressurization rates (Fig. 3k), while simultaneously reducing the amplitude of pressure oscillations (Fig. 3l). The reduction in pressure oscillations due to preheating supports the hypothesis that unsaturated thermodynamic conditions are the underlying cause of actuator vibrations. However, since higher standby pressures reduce the operating range, we settled on a standby pressure of 20 kPa as a compromise solution.

Additionally, we verified that operating under lower power settings also reduces pressure oscillations (Fig. 3m). Although a reduction of power would inevitably limit the actuator's pressurization and strain rate performance (Fig. 3n), this effect can be exploited to further reduce boiling instabilities during the initial preheating stage, without compromising performance under normal operating conditions.

The hardware implementation of the non-linear control strategy (Fig. 4a) is detailed in "Methods" section. The control strategy applies an ON-OFF controller to preheat actuators with a low constant heat flux of approximately 16 $kW/m^2$, until their standby pressure is reached. At this point, a PI controller takes over and is used to track pressure values within the actuator's operating range (Fig. 4b). The PI controller can impose higher heat fluxes of up to 64 $kW/m^2$ under normal operating conditions, a value that was empirically defined according to conservative safety requirements and the results of material testing available in the Supplementary Note 4. Results show the controller's ability to converge to either target pressures or displacements (Figs. 4c and 4d, and Supplementary Movie 2).

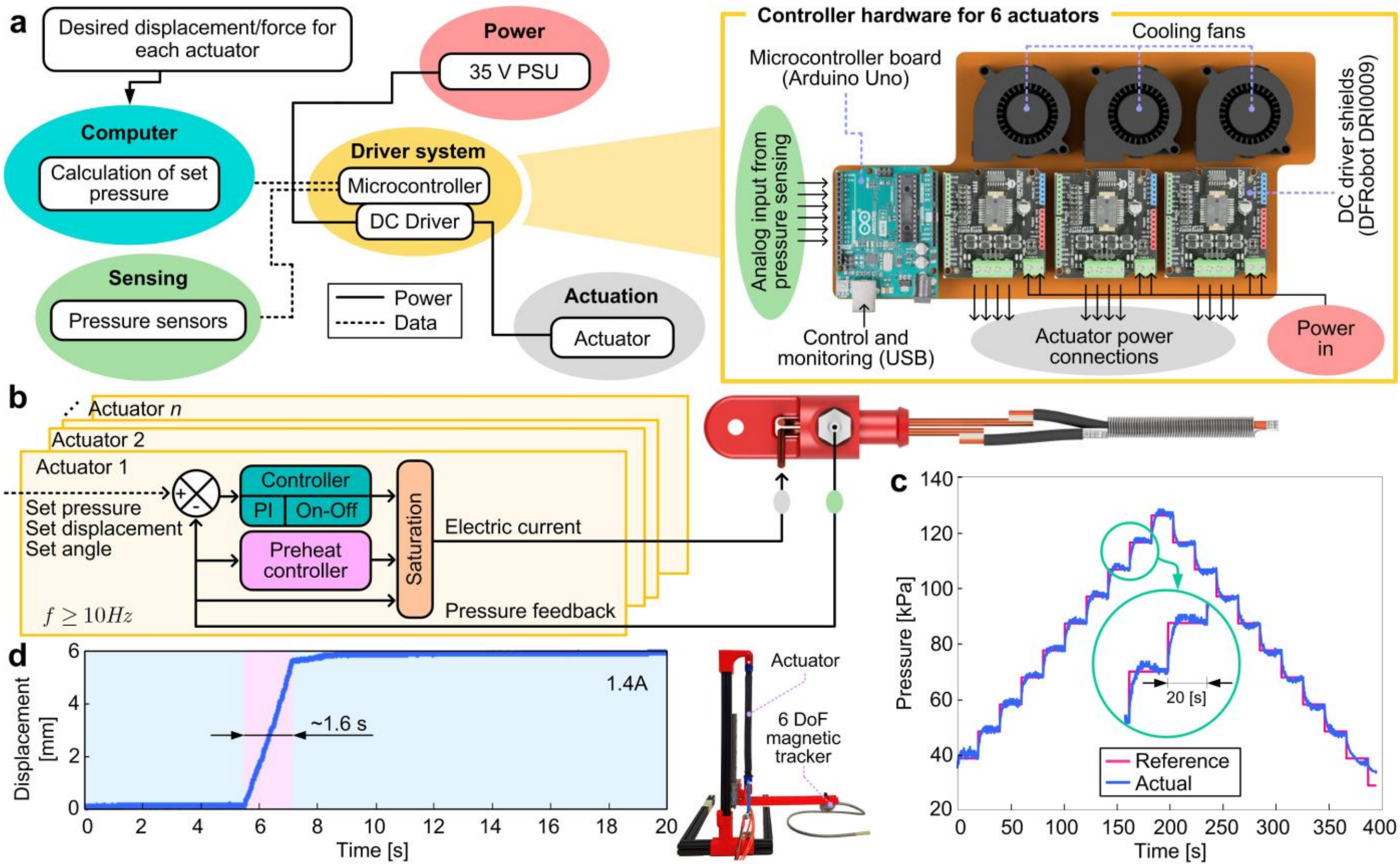

**Fig. 4. | Actuator control architecture and main elements. a** Controller hardware, system architecture and connection diagram. **b** Actuator controller based on PI or the On-Off method. **c** Control results for step target pressure (incremental steps of 10 kPa each lasting 20 s). The reduction in depressurization rates becomes apparent at pressures below 40 kPa. **d** The optimized PI-based actuator displacement controller takes ~1.6 s to reach a set displacement of 6 mm. Displacement values were obtained using a 6 Degrees of Freedom (DoF) magnetic tracker.

## Robot prototypes

The performance of our soft actuators combined with their compact form factor motivated us to explore how this technology can be integrated into different soft robots. We started by developing a biomimetic hand powered by three linear soft actuators (Fig. 5a), demonstrating how compliant soft actuators can be integrated into an otherwise non-compliant mechanical system. The actuators are linked to the finger joints using tendons made of nylon wire. A compact pulley system provides a 1:2 transmission ratio, duplicating the linear displacement provided by the actuators. A set of springs provides the tension required to return the actuators back to their extended position. Two actuators control the thumb and index fingers independently, while a third actuator controls the middle, ring and little fingers simultaneously. The fingers can close approximately five times faster than they can open (Fig. 5b), highlighting once again the limitation imposed by the rate of heat dissipation. The hand is capable of grasping objects of various shapes and sizes while benefiting from the compliance inherent to soft actuators (Fig. 5c and Supplementary Movie 3).

We then proceeded to demonstrate a soft gripping task by building an electric soft gripper powered by three bending actuators (Fig. 5d). Not accounting for the actuators themselves, the design and assembly of this gripper was completed in less than one hour, demonstrating the development efficiency enabled by general-use soft actuators. The gripper, attached to a robot, successfully performed pick-and-place tasks using fruit models of varying weights, sizes and geometries from the Yale-CMU-Berkeley (YCB) Object and Model Set (Figs. 5e and 5f, and Supplementary Video 4). Successful grasping of all models was achieved at pressures higher than 60 kPa (Fig. 5g).

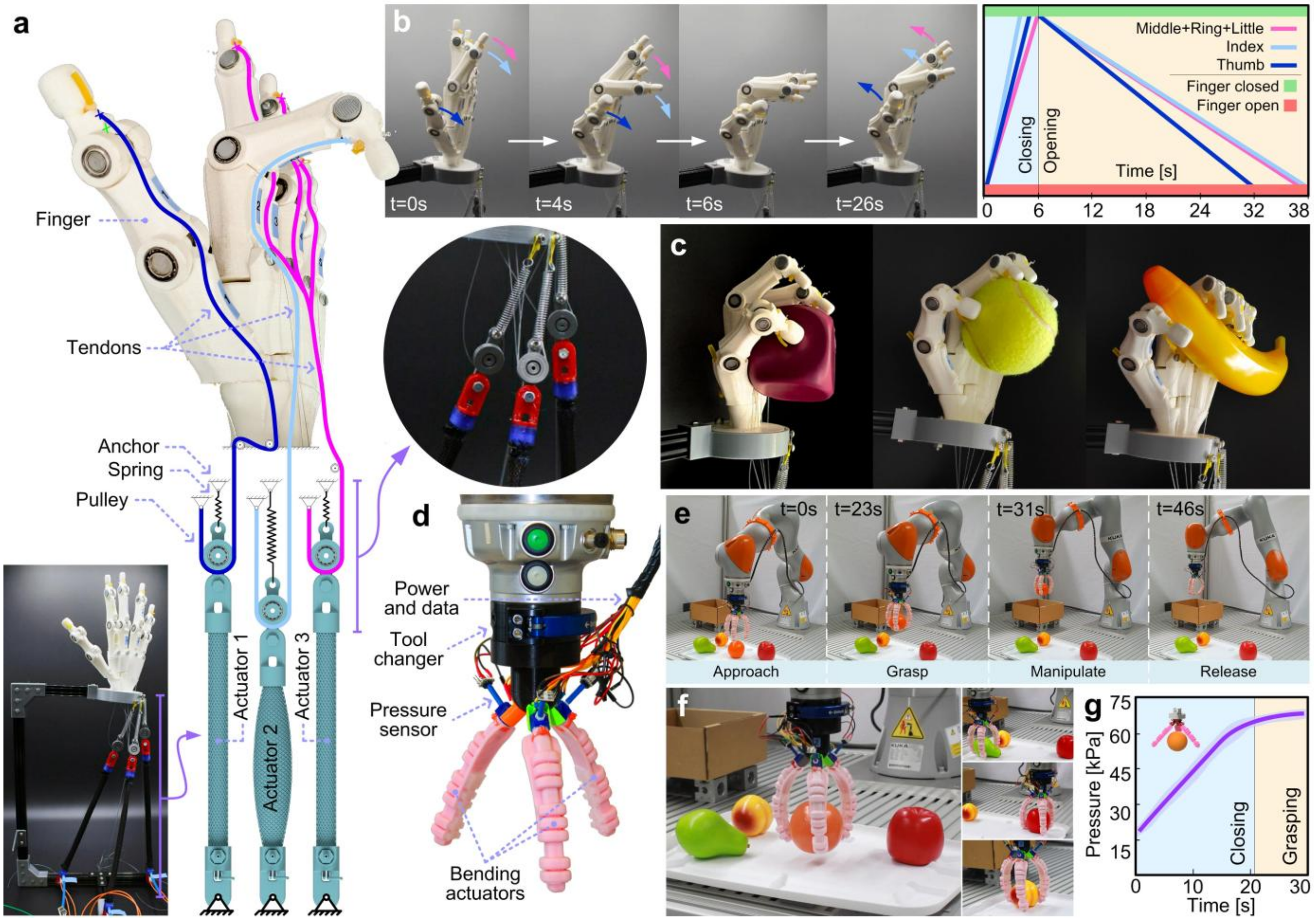

**Fig. 5. | Biomimetic hand and electric soft gripper. a** Tendon-driven biomimetic hand powered by three linear actuators using a spring pulley mechanism. The soft actuators provide compliant yet precise actuation. **b** The fingers' motion and the hand's opening/closing time individualized for each finger. **c** Biomimetic hand grasping various objects. **d** Soft gripper composed of three bending-type actuators attached to a collaborative robot flange. **e** Together, the robot and the gripper can perform pick-and-place operations. **f** The gripper can grasp and manipulate fruit models of varying weight and geometry. **g** Grasping an orange is achieved at pressures higher than 60 kPa and takes ~20 s with the actuators operating at 20W.

Finally, to demonstrate the potential of these actuators in a complex locomotion task, we designed and implemented Bixo, an electric soft quadruped robot powered by liquid-gas phase transition. Bixo is a development platform that can perform complex locomotion tasks such as climbing tubes and crawling along trees. We designed Bixo to enable rapid replacement of the mechatronic components, specifically the sliding structure and the five actuators. Four of these actuators are bending-type soft actuators, and the fifth is a DC motor with a reduction gear drive to actuate the sliding structure (SS) through a lead screw (Fig. 6a and Supplementary Note 8). Bixo's legs adjust to circular profiles, functioning as grippers, one for the back legs (BL) and another for the front legs (FL) (Fig. 6b). Locomotion on an inclined tube is achieved through a cyclic process in which (i) the robot's FL close to grasp the circular profile and fix the robot in place, (ii) the DC motor actuates the lead screw to position the open BL, (iii) the BL close to grasp the circular profile and fix the robot in place, (iv) the FL open to release the circular profile, and (v) the DC motor actuates the lead screw again to reposition the FL (Fig. 6c and Supplementary Movie 5). The first complete cycle of the actuation pattern was completed in ~80 s and after stabilization a faster cycle of ~25 s is achieved, considering a set pressure of ~60 kPa for both the BL and FL (Fig. 6d). This decrease in cycle rate is caused by the asymmetry between the pressurization and depressurization rates of the actuators, resulting in each subsequent cycle to begin at a higher pressure than the previous one, thereby requiring less time to reach the pressure needed for grasping the tube. This effect persists until the starting pressure of each cycle matches the maximum pressure at which the robot can release its grasp on the tube. At this point, the cycle rate stabilizes and ceases to decrease (Fig. 6d). Bixo demonstrated the capability to crawl on a horizontal tree trunk, showing compliance with the uncertainty of the rough surface of the trunk (Figs. 6e and 6f, and Supplementary Movie 5). This robotic platform can be equipped with multiple sensors and sets a solid foundation for the continuous development of untethered robots powered by liquid-gas phase transition actuators.

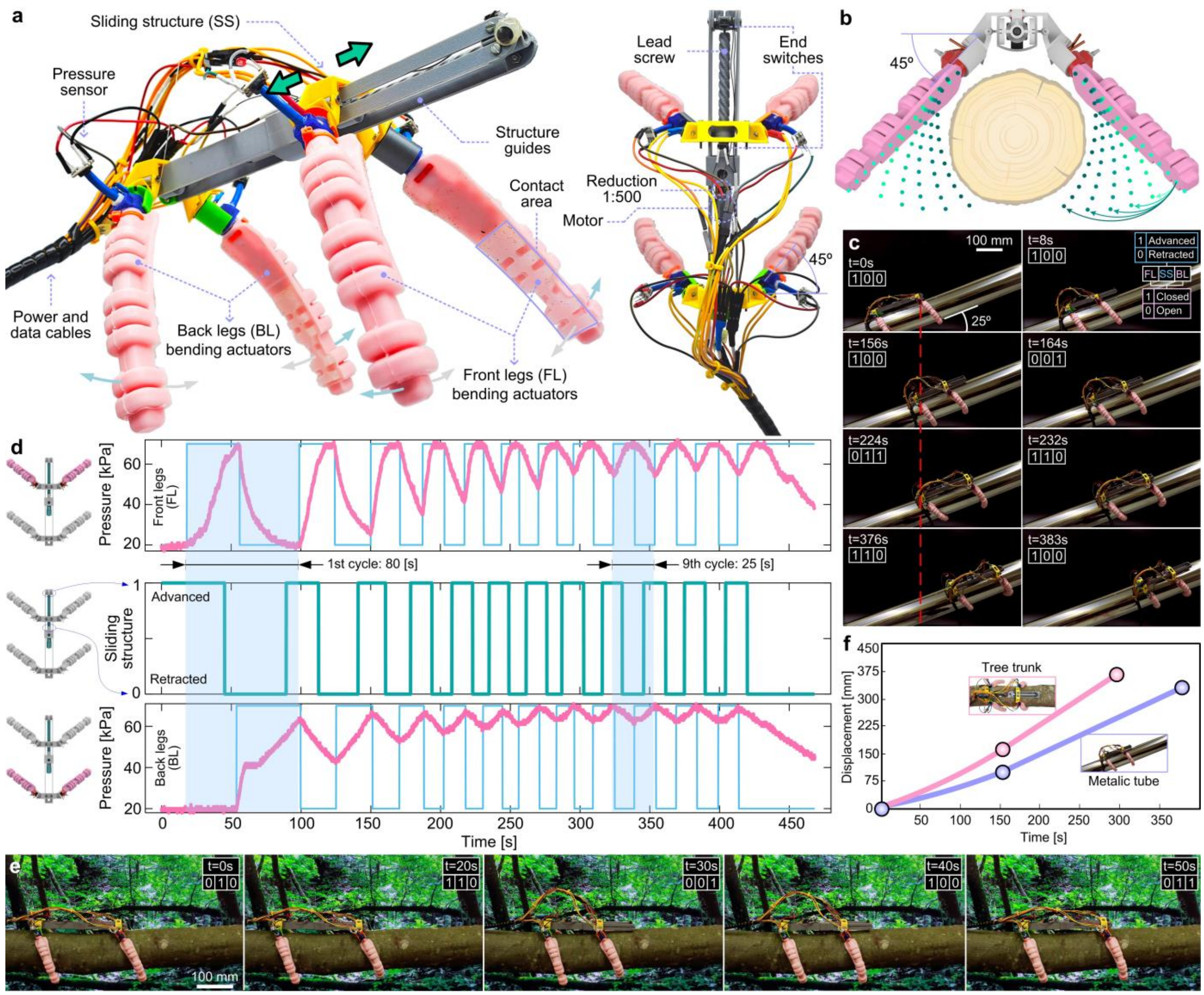

**Fig. 6. | Electric-driven soft quadruped robot Bixo. a** The Bixo design includes two bending-type actuators for each of the back and front legs, and a DC motor to actuate the lead screw that operates the sliding structure. **b** Robot legs assembled to grasp circular profiles. **c** Locomotion on an inclined tube. **d** Locomotion cyclic pattern considering a set pressure of ~60 kPa for both BL and FL. **e** Locomotion on an unstructured tree trunk. **f** Robot displacement in both environments.

## DISCUSSION

This work brings to light the hidden capabilities of high-power electric liquid-gas phase transition actuators. We introduce a soft actuator design capable of pressurization rates of over 100 kPa/s, and contractile strain rates of over 16%/s. We propose a comprehensive selection method for working fluids that allows actuators to meet specific application requirements. The instability responsible for actuator vibration at high power settings was also identified and addressed using a non-linear control approach.

The standardization of soft robotic technologies, including soft actuators, is a challenge that should be addressed in the coming years as an enabling strategy to increase the availability, functionality, and flexibility of increasingly complex robots. Standardizing electric soft actuators will be particularly challenging due to the difficulties of transmitting mechanical energy through soft materials. Actuators will need to be adaptable and capable of delivering mechanical energy in its final form without depending on the stiffness of traditional mechanical transmission mechanisms. Our actuator's modular design and physical adaptability facilitate their integration into innovative untethered robotic systems. Moreover, they are fabricated using off-the-shelf materials, electronics, and manufacturing equipment.

Regarding the limited depressurization rates observed throughout this work, our results suggest two possible strategies for future improvement. The first strategy is to increase the area to volume ratio of the actuators, through miniaturization. Heat energy stored in an actuator at a given temperature is proportional to its volume, while the rate of convective heat dissipation is proportional to its surface area. Accordingly, by increasing the area-to-volume ratio, a greater proportion of the total stored latent heat energy could be dissipated during the same time period. A second strategy for improving depressurization rates would be to shift the entire operating pressure range towards higher values, simultaneously increasing both the preheated standby and maximum operating pressures, at the cost of a reduced factor of safety. This strategy is based on our observation that higher pressures lead to higher depressurization rates. However, operating at higher pressures without reducing the actuator's displacement range would require the use of preloading mechanisms, adding complexity to the design. Additionally, little is known about the life expectancy of liquid-gas phase transition actuators. During this work, the actuators endured over a $10^3$ actuation cycles over 10 months of testing, but their actual life expectancy and modes of failure remain to be studied.

Without overlooking the limitations in efficiency imposed by electrothermal soft actuation, we argue that liquid-gas soft actuators offer a combination of performance and adaptability that should not be ignored and could potentially result in significant gains in system-wide efficiency. An example of this can be seen in Bixo, which required an average of 50 W of power across all actuators during its tube-climbing task. Although this value is difficult to benchmark due to the unique nature of the robot, it is plausible to argue that it is within the same order of magnitude as what would be expected of a similarly sized non-soft robot, performing a similar task. This is a key contribution towards reaching the inflection point where soft robots become an enticing alternative to rigid robotic systems, moving them from the laboratory into our daily lives.

## METHODS

**Materials**: The actuator end terminals were 3D printed in thermoplastic polyurethane (TPU 98A, RS PRO, United Kingdom) due to its high softening point of 138 ℃. The heating element coil is made of FeCrAl wire (Kanthal A, Kanthal, Sweden) due to its higher electrical resistivity when compared to other common alloys such as NiCr, NiFe or CuNi. This enables the actuators to operate with lower currents, thereby reducing electrical losses. While NiCr alloys present better wet corrosion resistance, the aluminum oxide ($Al_2O_3$) formed on FeCrAl has better adhesion to the alloy's surface and it is a better electrical insulator when compared to chromium oxide ($Cr_2O_3$) formed on NiCr. The actuator's soft structures were fabricated using platinum-cured silicones. Unless specified otherwise, linear actuators were made with Ecoflex 00-30 (Smooth-On, USA). Bending actuators were fabricated using Smooth-Sil 940 (Smooth-On, USA), with an additional layer of Ecoflex 00-30 to increase friction, which is required for grasping applications. The linear actuator's braided sleeve is made of Polyethylene Terephthalate (PET) (article. no. 06240405010, SES Sterling, France). The rigid structural elements of the robot prototypes were 3D printed using acrylonitrile butadiene styrene (ABS, RS PRO, United Kingdom).

**Fabrication of actuator's heating element**: The heating element coil is wound around a high-temperature silicone-insulated electric wire and connected to copper terminals on one side using an electrical wire ferrule. The high temperature wire is used to connect the opposite end of the coil. Copper terminals are passed through the TPU structural terminals and sealed using cyanoacrylate adhesive (401, Loctite, Germany). Details of the fabrication process are available in the Supplementary Note 2.

**Fabrication of the actuator's soft structures**: The linear actuator's soft structure is a cylindrical tube, 170 mm long, with an internal diameter of 6 mm and an external diameter of 9 mm. It was cast using 3D printed molds and aluminum cores. The tube is then sheathed with the braided sleeve. The geometric parameters of the braid, which influence the theoretical maximum actuator displacement, are as detailed in Supplementary Note 5. The bending actuator variant features a complex-shaped hollow soft structure, which is cast using 3D printed molds and a water-soluble hollow core. The core is 3D printed in polyvinyl alcohol (Smartfil PVA, Smart Materials 3D, Spain) and later dissolved in a water bath at 60 ℃. To expedite the dissolution process, heated water is circulated under pressure through the hollow core using water pumps [10]. The molds, water-soluble cores, and robot components were 3D printed on a commercially available FFF 3-axis single-nozzle machine (Prusa i3 MK3S+, Prusa, Czechia). The detailed fabrication process of both the linear and bending actuator variants is detailed in Supplementary Note 3.

**Control and monitoring elements**: An actuator control unit is composed of an off-the-shelf microcontroller (Arduino Uno, Arduino, Italy) and DC motor driver shields based on the L298n full-bridge driver (DRI0009, DFRobot, China) (Fig. 4a). The controller can drive up to 6 actuators simultaneously, at currents of up to 2A each. A high-power version of the controller, used for the isometric shock test, was also assembled by replacing the DC motor shields with 40 A DC solid-state relays. A pressure sensor is installed in each actuator to provide pressure feedback (ABPDANV060PGAA5, Honeywell, USA). The microcontroller receives feedback from the sensors and sends commands to the actuator while running the proposed controllers. The graphical user interface (GUI) and wiring diagram are detailed in Supplementary Figures 14 and 15.

**Characterization experiments**: Isometric and isotonic experimental tests were conducted at room temperature (20 to 25 °C, 1 atm). Force values were acquired using a 5 kg capacity load cell (TAL220B, HT Sensor Technology Co, China). Internal actuator temperature was monitored using a digital temperature sensor (DS18B20, Maxim Integrated, USA) installed in the actuators solely for characterization purposes. This sensor is not required during normal operation of the actuators. Displacement feedback was captured using a 6 DoF magnetic tracker (Liberty, Polhemus, USA) and bending angles were measured using a custom MATLAB video processing program. Data analysis was conducted using a custom python interface that reads data from the microcontroller. A validated static model, which associates pressures, displacements, and static forces, was developed using data from isotonic tests, detailed in Supplementary Note 6.

**Data availability:** All data supporting the findings and conclusions of this study are provided in the main text and supplementary materials. Data necessary for replicating the actuators and robot prototypes are also included. For additional information, please contact the corresponding author. Updated drawings and computer code are available in the GitHub repository https://github.com/softrobotic/bixo, and the Zenodo repository https://doi.org/10.5281/zenodo.15013682.

**Acknowledgments:** We thank A. Silva and P. Matos for the valuable discussions and suggestions on the fabrication of soft actuators. We also thank P. Costa for the technical support in the realization of the robot's experimental tests. This work was supported by the Portuguese Foundation for Science and Technology grant UIDB/00285/2020 (P.N.), LA/P/0112/2020 (P.N.), and 2022.13512.BD (D.F.).

**Author Contributions:**

Conceptualization: D.F., P.N. Methodology: D.F., P.N. Investigation: D.F. Visualization: D.F., P.N. Funding acquisition, project administration and supervision: P.N. Writing – original draft: D.F. Writing – review and editing: P.N.

**Competing Interests:** Authors declare that they have no competing interests.

**Additional Information:** Supplementary information.

Supplementary Information

# Electrically-driven phase transition actuators to power soft robot designs

D. Fonseca,[1] P. Neto[1]*

[1] *University of Coimbra, CEMMPRE, Department of Mechanical Engineering, Coimbra, Portugal*

*Corresponding author: Pedro Neto, pedro.neto@dem.uc.pt*

**Supplementary Information file includes:**

Supplementary Notes 1 to 8

Supplementary Figures 1 to 15

Supplementary Tables 1 to 17

## Supplementary Note 1: CALCULATION OF THE BOUNDARY WORK RATIO

### Enthalpy of vaporization across a range of saturation temperatures

The estimation of a fluid's enthalpy of vaporization $\Delta H_{vap}$ at different saturation temperatures from a single known value pair can be made using Watson's widely accepted method[73]:

$$\frac{\Delta H_1}{\Delta H_2} = \left( \frac{1 - T_{r1}}{1 - T_{r2}} \right)^{0.38} \tag{1}$$

Where $\Delta H_n$ [J/kg] is the enthalpy of vaporization at saturated reduced temperature $T_{rn}$. Reduced temperatures are a dimensionless thermodynamic coordinate, defined by:

$$T_r = \frac{T}{T_c} \tag{2}$$

Here, $T$ [K] is the fluid's actual absolute temperature and $T_c$ [K] is the fluid's critical temperature. The reference values for all analyzed working fluids are summarized in Supplementary Table 1. From Supplementary Equation (2) and Supplementary Table 1, we can rearrange Supplementary Equation (1) to approximate the enthalpy of vaporization over a range of relevant saturation temperatures, as plotted in Supplementary Fig. 1.

### Pressure-temperature saturation curves

Pressure values corresponding to a given saturation temperature $T_{sat}$ can be calculated from the fluid's pressure-temperature saturation curve. The equations defining this curve for all analyzed working fluids are summarized in Supplementary Tables 2 and 3, and plotted in Supplementary Figure 2.

**Supplementary Table 1**. Working fluids physical properties.

| FLUID | $T_{sat}$ @1 $atm$ $[K]$ | $T_c$ $[K]$ | $\Delta H_{vap}$ @1 $atm$ $[J / kg]$ | MOLAR WEIGHT $[kg / mol]$ |
|---|---|---|---|---|
| WATER | 373.15 | 647.00 | 2 256 400.00 | 1.80153E-02 |
| FC-72 | 329.15 | 449.15 | 88 000.00 | 3.38000E-01 |
| PF-5052 | 323.15 | 454.00 | 105 000.00 | 2.99000E-01 |
| PF-5060 | 329.15 | 451.00 | 88 000.00 | 3.38000E-01 |
| PF-5070 | 353.15 | 478.00 | 80 000.00 | 3.88000E-01 |
| NOVEC 7000 | 307.15 | 438.15 | 142 000.00 | 2.00000E-01 |
| NOVEC 7100 | 334.15 | 468.15 | 112 000.00 | 2.50000E-01 |
| NOVEC 7200 | 349.15 | 483.15 | 119 000.00 | 2.64000E-01 |
| ETHANOL | 351.50 | 514.00 | 837 016.00 | 4.60684E-02 |
| ACETONE | 329.30 | 508.00 | 501 040.00 | 5.80791E-02 |
| R-11 | 296.90 | 471.15 | 181 360.00 | 1.37368E-01 |
| PENTANE | 309.20 | 469.80 | 357 706.60 | 7.21488E-02 |
| METHANOL | 337.80 | 513.00 | 1 101 096.00 | 3.20419E-02 |
| HEXANE | 341.90 | 507.60 | 335 136.11 | 8.61754E-02 |
| BENZENE | 353.30 | 562.00 | 393 696.94 | 7.81118E-02 |
| HEPTANE | 371.50 | 540.00 | 316 685.70 | 1.00202E-01 |
| TOULENE | 383.80 | 593.00 | 360 696.52 | 9.21384E-02 |

**Supplementary Table 2**. Vapor pressure of working fluids.

| FLUID | EQUATION OF SATURATION CURVE | UNITS | SOURCE |
|---|---|---|---|
| WATER | $P_{sat}(T) = 0.61121 \times \exp\left(\left(18.678 - \frac{T}{234.5}\right) \times \left(\frac{T}{257.14 + T}\right)\right)$ | [kPa], [ºC] | Ref. 74 |
| FC-72 | $P_{sat}(T) = 10^{9.729 - \frac{1562}{T}}$ | [Pa], [K] | 3M |
| PF-5052 | $P_{sat}(T) = 10^{10.062 - \frac{1643}{T}}$ | [Pa], [K] | 3M |
| PF-5060 | $P_{sat}(T) = 10^{9.729 - \frac{1562}{T}}$ | [Pa], [K] | 3M |
| PF-5070 | $P_{sat}(T) = 10^{10.142 - \frac{1824}{T}}$ | [Pa], [K] | 3M |
| NOVEC 7000 | $P_{sat}(T) = \exp\left(\frac{-3548.6}{T} + 22.978\right)$ | [Pa], [K] | 3M |
| NOVEC 7100 | $P_{sat}(T) = \exp\left(22.415 - 3641.9 \times \frac{1}{T}\right)$ | [Pa], [K] | 3M |
| NOVEC 7200 | $P_{sat}(T) = \exp\left(22.289 - 3752.1 \times \frac{1}{T}\right)$ | [Pa], [K] | 3M |
| R-11 | $P_{sat}(T) = 10^{\left(4.01447 - \frac{1043.303}{T - 36.602}\right)}$ | [bar], [K] | NIST |
| PENTANE | $P_{sat}(T) = 10^{\left(3.9892 - \frac{1070.617}{T - 40.454}\right)}$ | [bar], [K] | NIST |
| METHANOL | $P_{sat}(T) = 10^{\left(5.20409 - \frac{1581.341}{T - 33.5}\right)}$ | [bar], [K] | NIST |
| HEXANE | $P_{sat}(T) = 10^{\left(4.00266 - \frac{1171.53}{T - 48.784}\right)}$ | [bar], [K] | NIST |
| BENZENE | $P_{sat}(T) = 10^{\left(4.72583 - \frac{1660.652}{T - 1.461}\right)}$ | [bar], [K] | NIST |
| HEPTANE | $P_{sat}(T) = 10^{\left(4.02832 - \frac{1268.636}{T - 56.199}\right)}$ | [bar], [K] | NIST |
| TOULENE | $P_{sat}(T) = 10^{\left(4.07827 - \frac{1343.943}{T - 53.773}\right)}$ | [bar], [K] | NIST |

**Supplementary Table 3.** Vapor pressure of acetone and ethanol[75].

| FLUID | $\log_{10}(P[mmHg]) = A + \frac{B}{T[K]} + C \times \log_{10}(T[K]) + D \times T[K] + E \times T[K]^2$ | | | | |
|---|---|---|---|---|---|
| | A | B | C | D | E |
| Acetone | 28.5884 | –2.4690E+03 | –7.3510E+00 | 2.8025E–10 | 2.7361E–06 |
| Ethanol | 23.8442 | –2.8642E+03 | –5.0474E+00 | 3.7448E–11 | 2.7361E–07 |

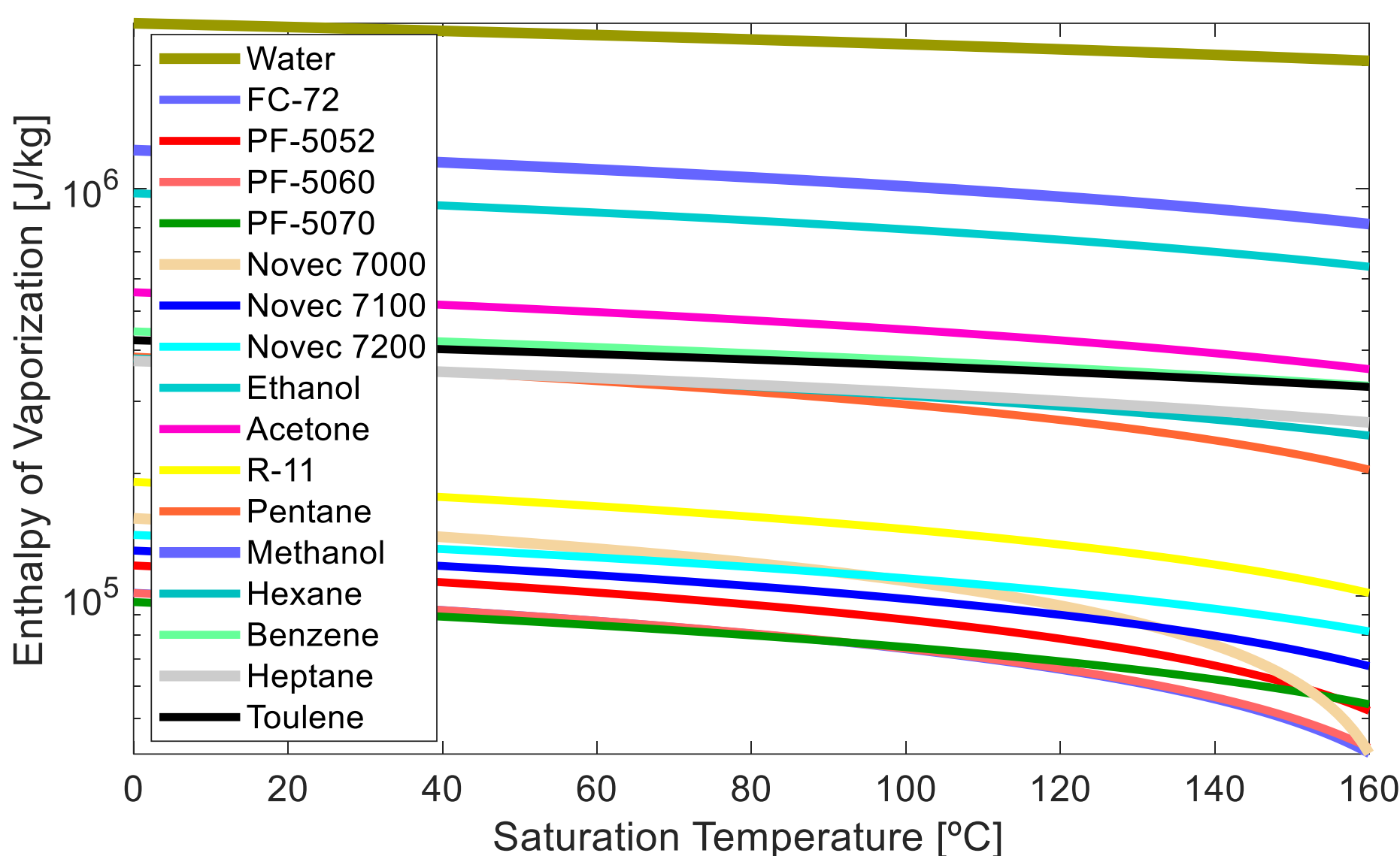

**Supplementary Fig. 1. Fluid's enthalpy of vaporization**.

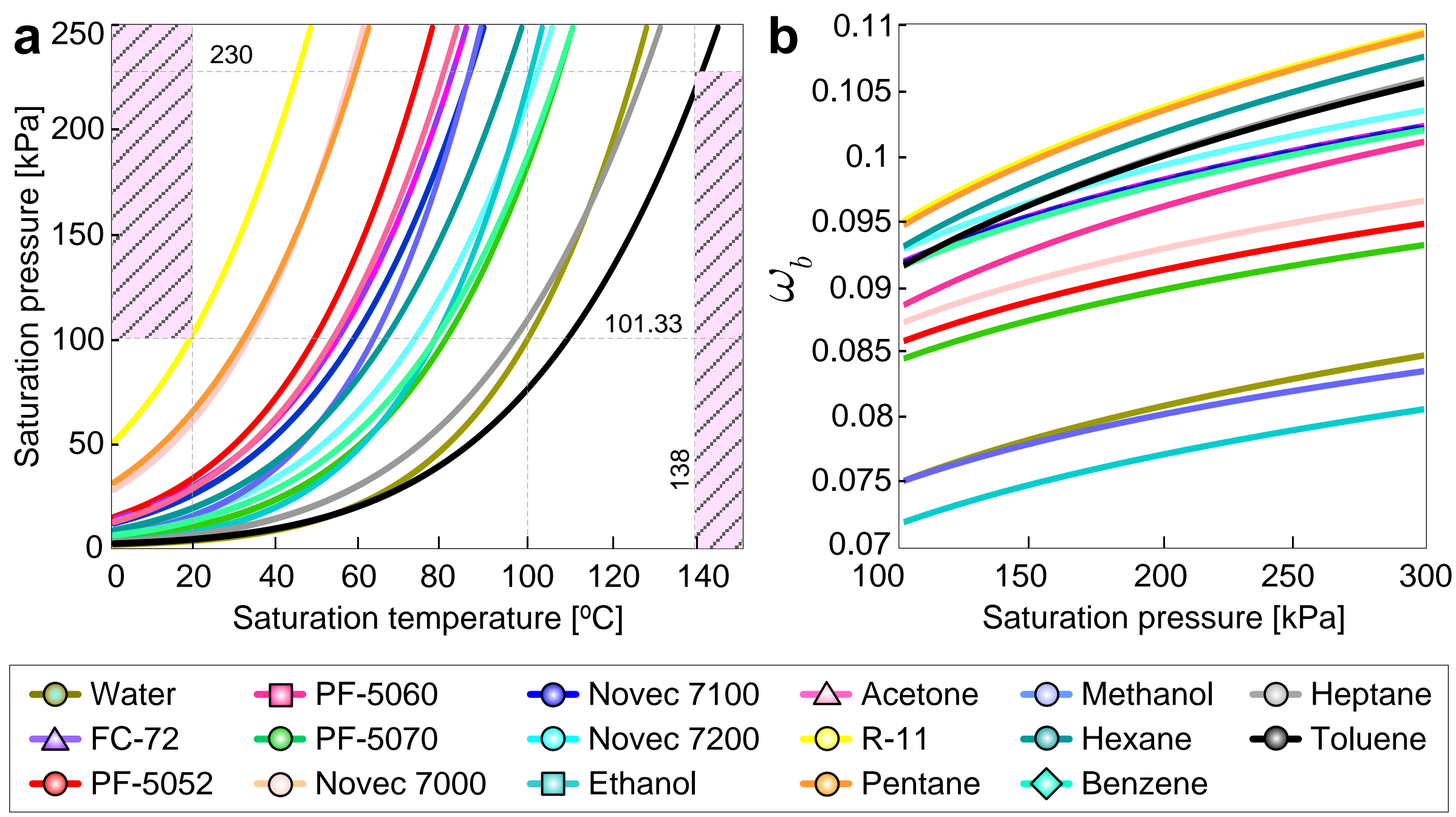

**Supplementary Fig. 2. Fluid's saturation curves**. **a** Saturation pressure relative to the saturation temperatures. **b** Boundary work ratio relative to the saturation pressure.

**Volumetric expansion**

The work done by the expanding fluid under isobaric conditions can be determined by:

$$W = -P\Delta V \tag{3}$$

Where $\Delta V$ is the total change in volume and $P$ is the pressure at which the expansion took place. We consider energy to be negative when leaving the fluid, which is why there is a minus sign in Supplementary Equation (3). The enthalpy of vaporization accompanying a phase change is related to other system properties:

$$\Delta H_{vap} = T\Delta V \frac{dP_{sat}}{dT} \tag{4}$$

The previous equation may be rearranged to calculate the total volumetric change:

$$\Delta V = \frac{\Delta H_{vap}}{T}\left(\frac{dP_{sat}}{dT}\right)^{-1} \tag{5}$$

The pressure slope $dP_{sat}/dT$ can be calculated from the saturation curve equations in Supplementary Table 2 and $\Delta H_{vap}$ was previously determined through Supplementary Equation (1), so we may proceed to calculate the effective volumetric change, $\Delta V$, at each point in the fluid's saturation curve.

**Boundary work ratio**

By definition, the enthalpy of a system is explicitly defined by:

$$H = U + PV \tag{6}$$

Where $U$ is the system's internal energy. Since $U$, $P$ and $V$ are state functions, the enthalpy $H$ must also be a state function, and thus, any finite change in its value can be determined by:

$$\Delta H = \Delta U + \Delta(PV) \tag{7}$$

Which, assuming isobaric expansion becomes:

$$\Delta H = \Delta U + P\Delta(V) \quad (8)$$

Remembering Supplementary Equation (3) leads us to:

$$\Delta H = \Delta U - W = Q + W - W = Q \quad (9)$$

Where $Q$ represents the heat energy exchanged through the boundaries of the system. Finally, the boundary work ratio, can be calculated through:

$$\omega_b = \left|\frac{W}{Q}\right| = \left|\frac{-P\Delta V}{\Delta H_{vap}}\right| \quad (10)$$

**Supplementary Table 4**. Working fluids symptoms of exposure.

| FLUID | SYMPTOMS OF EXPOSURE | | | | | |
|---|---|---|---|---|---|---|
| | Skin | Eye | Oral | Carcinogenicity | Reproductive Toxicity | Source |
| Benzene | Irritating | Serious irritation | Irritating | Possible | - | Sigma-Aldrich |
| Heptane | Irritating | No irritation | May be fatal | - | - | Sigma-Aldrich |
| Toulene | Irritating | No irritation | - | - | Suspected | Sigma-Aldrich |
| Hexane | Irritating | No irritation | - | - | Suspected | Sigma-Aldrich |
| Methanol | No irritation | No irritation | Irritation | Possible | - | Sigma-Aldrich |
| Pentane | No irritation | No irritation | May be fatal | - | - | Sigma-Aldrich |
| R-11 | Irritating | Irritation | Irritating | - | - | National Refrigerants, inc. |
| Acetone | Mild irritation | Irritation | - | - | - | Sigma-Aldrich |
| Ethanol | No irritation | Serious irritation | - | - | - | Sigma-Aldrich |
| Novec 7200 | No irritation | No irritation | May be harmful if swallowed | - | - | 3M |
| Novec 7100 | No irritation | No irritation | - | - | - | 3M |
| Novec 7000 | No irritation | No irritation | - | - | - | 3M |
| PF-5070 | Minimally irritating | No irritation | Not irritating | - | - | 3M |
| PF-5060 | No irritation | No irritation | - | - | - | 3M |
| PF-5052 | No irritation | No irritation | - | - | - | 3M |
| FC-72 | No irritation | No irritation | - | - | - | 3M |
| Water | No irritation | No irritation | Not irritating | Not expected | Not toxic | Sigma-Aldrich |

**Supplementary Table 5**. Working fluids acute toxicity.

| FLUID | ACUTE TOXICITY | | | | | | | | | |
|---|---|---|---|---|---|---|---|---|---|---|
| | $LD_{50}$ (Oral) | | | $LC_{50}$ (Rat, 4h) | | | $LD_{50}$ (Dermal) | | | Source |
| | Value | Units | Remarks | Value | Units | Remarks | Value | Units | Remarks | |
| Benzene | 3002 | [mg/kg] | OECD Test Guideline 401 | - | [mg/l] | - | 13630 | [mg/kg] | Rabbit (iuclid) | Sigma-Aldrich |
| Heptane | >5000 | [mg/kg] | OECD Test Guideline 401 | >29.29 | [mg/l] | OECD Test Guideline 403 | >2000 | [mg/kg] | OECD Test Guideline 402 | Sigma-Aldrich |
| Toulene | 5580 | [mg/kg] | Directive 67/548/EEC, Annex V, B.1. | 25.7 | [mg/l] | OECD Test Guideline 403 | >5000 | [mg/kg] | Rabbit (ECHA) | Sigma-Aldrich |
| Hexane | 16000 | [mg/kg] | OECD Test Guideline 401 | 172 | [mg/l] | RTECS | >2000 | [mg/kg] | OECD Test Guideline 402 | Sigma-Aldrich |
| Methanol | 2769 | [mg/kg] | - | 128.2 | [mg/l] | - | 17100 | [mg/kg] | - | Sigma-Aldrich |
| Pentane | >2000 | [mg/kg] | OECD Test Guideline 401 | >25.3 | [mg/l] | OECD Test Guideline 403 | - | - | - | Sigma-Aldrich |
| R-11 | - | - | 26000 | [ppm] | - | - | - | - | - | National Refrigerants, inc. |
| Acetone | 5800 | [mg/kg] | ECHA | 76 | [mg/l] | - | 20000 | [mg/kg] | Rabbit (iuclid) | Sigma-Aldrich |
| Ethanol | 10470 | [mg/kg] | OECD Test Guideline 401 | 124.7 | [mg/l] | OECD Test Guideline 403 | - | - | - | Sigma-Aldrich |
| Novec 7200 | >2000 | [mg/kg] | - | >989 | [mg/l] | - | 2000 | [mg/kg] | - | 3M |
| Novec 7100 | >5000 | [mg/kg] | - | >1000 | [mg/l] | - | >5000 | [mg/kg] | - | 3M |
| Novec 7000 | >2000 | [mg/kg] | - | 820 | [mg/l] | - | >5000 | [mg/kg] | - | 3M |
| PF-5070 | NA | - | - | - | - | - | NA | - | - | 3M |
| PF-5060 | >5000 | [mg/kg] | - | >276 | [mg/l] | - | >5000 | [mg/kg] | - | 3M |
| PF-5052 | >5000 | [mg/kg] | - | >15.4 | [mg/l] | - | >5000 | [mg/kg] | - | 3M |
| FC-72 | >5000 | [mg/kg] | - | >276 | [mg/l] | - | >5000 | [mg/kg] | - | 3M |

| Water | >90000 | [mg/kg] | RTECS | - | - | - | - | - | - | Sigma-Aldrich |
|---|---|---|---|---|---|---|---|---|---|---|

**Supplementary Table 6**. Working fluids ecotoxicological information.

| FLUID | ECOTOXICOLOGICAL DATA | | | | | | |
|---|---|---|---|---|---|---|---|
| | Ozone depleting | Remarks | Global warming potential | Remarks | Marine pollutant | Remarks | Source |
| Benzene | Not expected | - | - | - | Very toxic to aquatic life | H410 | Sigma-Aldrich |
| Heptane | Not expected | - | - | - | Very toxic to aquatic life | H400, H410 | Sigma-Aldrich |
| Toulene | Not expected | - | - | - | Harmful to aquatic life | H412 | Sigma-Aldrich |
| Hexane | Not expected | - | - | - | Toxic to aquatic life | H411 | Sigma-Aldrich |
| Methanol | Not expected | - | - | - | Not expected | - | Sigma-Aldrich |
| Pentane | Not expected | - | - | - | Toxic to aquatic life | H411 | Sigma-Aldrich |
| R-11 | Class 1 ODS, H420 | US EPA, GHS | - | - | Not expected | - | National Refrigerants, inc. |
| Acetone | Not expected | - | - | - | Not expected | - | Sigma-Aldrich |
| Ethanol | Not expected | - | - | - | Not expected | - | Sigma-Aldrich |
| Novec 7200 | - | - | - | - | - | - | 3M |
| Novec 7100 | - | - | - | - | - | - | 3M |
| Novec 7000 | - | - | - | - | - | - | 3M |
| PF-5070 | Not expected | - | High potential | - | - | - | 3M |
| PF-5060 | - | - | - | - | Hazardous to the aquatic environment | H413 | 3M |
| PF-5052 | - | - | - | - | Hazardous to the aquatic environment | H413 | 3M |
| FC-72 | - | - | - | - | - | - | 3M |
| Water | Not expected | - | - | - | No | - | Sigma-Aldrich |

**Supplementary Table 7**. Working fluids physical characteristics.

| FLUID | PHYSICAL DATA | | | |
|---|---|---|---|---|
| | Boiling Point @ 1 atm [°C] | Flash Point [°C] | Autoignition Temperature [°C] | Source |
| Benzene | 80.1 | -11 | 498 | Sigma-Aldrich |
| Heptane | 98 | -4 | 223 | Sigma-Aldrich |
| Toulene | 110 | 4.4 | - | Sigma-Aldrich |
| Hexane | 69 | -22 | 225 | Sigma-Aldrich |
| Methanol | 64.7 | 9.7 | 455 | Sigma-Aldrich |
| Pentane | 36.1 | -40 | 260 | Sigma-Aldrich |
| R-11 | 23.6 | No flash point | - | National Refrigerants, inc. |
| Acetone | 56 | -17 | 465 | Sigma-Aldrich |
| Ethanol | 78 | 13 | 455 | Sigma-Aldrich |
| Novec 7200 | 76 | No flash point | 375 | 3M |
| Novec 7100 | 61 | No flash point | 405 | 3M |
| Novec 7000 | 34 | No flash point | 415 | 3M |
| PF-5070 | 80 | - | - | 3M |
| PF-5060 | 56 | No flash point | - | 3M |
| PF-5052 | 50 | No flash point | - | 3M |
| FC-72 | 56 | No flash point | - | 3M |
| Water | 100 | No flash point | - | Sigma-Aldrich |

**Supplementary Note 2: FABRICATION AND CHARACTERIZATION OF THE HEATING ELEMENT**

The fabrication of the heating element follows a step-by-step procedure, Supplementary Table 8 and Supplementary Fig. 3. The heating element's main component is an electrically resistive coil, which can be dimensioned to meet distinct optimization requirements. In this work, we used two coil variants, differing only in length. The first variant was dimensioned to operate at a specific voltage and current, namely 1.7 A at 12 V. The second variant was tailored for maximum coil surface area, increasing the heat transfer rate and subsequently the strain rate performance of the actuator. The following subsections will provide an in-depth explanation of the two dimensioning strategies we followed, detailing the design considerations and fabrication techniques employed for each variant.

**Supplementary Table 8**. Fabrication of the actuator's heating element.

| FABRICATION STEP | DESCRIPTION | FIGURE(S) |
|---|---|---|
| 1 | 3D-print the component part file "linear_pressure_cap.stl" using a thermally resistant material, TPU 98A. | Supplementary Fig. 3a |
| 2 | Cut two pieces of copper wire, each measuring no less than 7.5 cm in length. | Supplementary Fig. 3b |
| 3 | Solder a 0.75 mm$^2$ wire ferrule, with the plastic insulation removed to the end of each copper wire. | Supplementary Fig. 3c and 3d |
| 4 | Cut one piece of 0.75 mm$^2$ silicone insulated electric cable with a length one centimeter longer than the desired heating coil length.<br>Strip 5 mm of insulation on both ends of the cable. | Supplementary Fig. 3e |
| 5 | Cut the heating element coil to the desired length. Stretch the coil to achieve a pitch that is at least twice the wire diameter. | Supplementary Fig. 3f |
| 6 | Crimp one end of the silicone cable prepared in Step 4 to one of the ferrules soldered on Step 3. | Supplementary Fig. 3g |
| 7 | Crimp one end of the coil prepared in Step 5 to the second ferrule soldered on Step 3. | Supplementary Fig. 3h |
| 8 | Pass the silicone cable through the inside of the coil. Then, use one additional wire ferrule to join the unconnected end of the cable to the coil. | Supplementary Fig. 3i |
| 9 | Insulate all wire ferrules using heat shrinking tubes. Alternatively, a thin layer of epoxy resin can be used to provide electric insulation. | Supplementary Fig. 3j |
| 10 | Insert the unconnected ends of the copper wires through the round holes in the pressure cap. | Supplementary Fig. 3k |
| 11 | Find the small depression on the backside of the pressure cap and fill it with cyanoacrylate glue to help seal the surrounding area. | Supplementary Fig. 3l |

| 12 | Insert a 4 mm silicone pneumatic tube through the pressure tap hole on the side of the pressure cap. Use cyanoacrylate glue to seal the joint, Supplementary Fig. 3m. Alternatively, a barbed fluid connector can be used, Supplementary Fig. 3n. | Supplementary Fig. 3m and 3n |
|---|---|---|

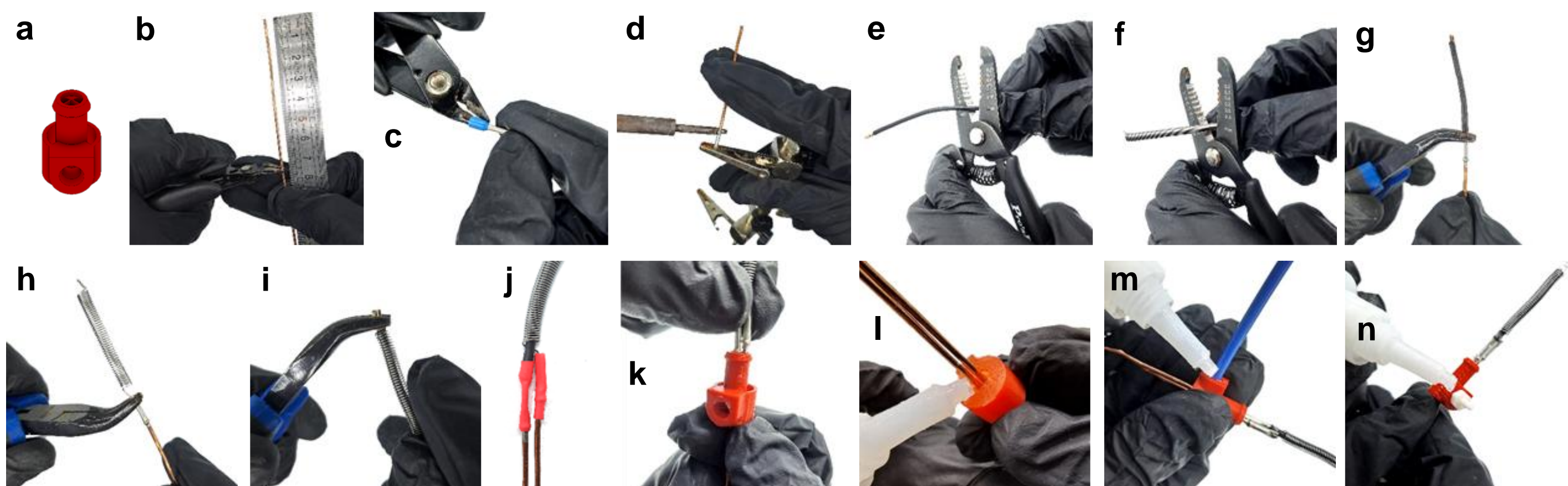

**Supplementary Fig. 3. Fabrication steps of the heating element**. **a** Fabrication step 1. **b** Fabrication step 2. **c** Fabrication step 3. **d** Fabrication step 3. **e** Fabrication step 4. **f** Fabrication step 5. **g** Fabrication step 6. **h** Fabrication step 7. **i** Fabrication step 8. **j** Fabrication step 9. **k** Fabrication step 10. **l** Fabrication step 11. **m** Fabrication step 12. **n** Fabrication step 12.

### Coil wire diameter

The wire diameter has a direct influence on both the coil's surface area and electrical resistance. Increasing the coil's surface area allows for either an increase in the rate of heat transfer or a reduction in heat flux, thereby reducing thermal stress on the system. If the coil is designed to fit within a particular volume, smaller diameter wires are usually preferable as they increase the surface area of the coil. This means that smaller diameters allow for smaller coil pitches, more turns, and longer wire lengths, thereby increasing the coil's surface area. An example of this is shown in Supplementary Fig. 4a, which displays how the surface area varies with different wire diameters, in this case assuming a constant coil diameter of 4 mm and a coil length of 10 mm, with a pitch equal to each wire diameter. Here, smaller wire diameters also result in increased electrical resistance, Supplementary Fig. 4b. The material resistivity used for the calculation is $\rho = 1.45\times10^{-6}$ Ω.m, the same as the FeCrAl alloy wires used throughout this work.

An exception occurs when a coil is dimensioned to meet a specific electrical requirement, and the material's resistivity cannot be changed. In that case, selecting wires with larger diameters will increase the length of wire required to achieve the desired electrical resistance value, resulting in a coil with higher surface area. Increasing the wire diameter results in a larger cross-sectional area and a lower electrical resistance per unit length. Consequently, a longer wire length is required to achieve the desired resistance, thus increasing the coil's surface area. An example of this can be seen in Supplementary Fig. 4c, where the surface area of a 1 Ω coil is calculated for a range of wire diameters.

As this work focuses on achieving high rates of heat transfer, which is essential for high actuator performance, smaller diameter wires are preferable. All the coils manufactured throughout this work used a wire diameter of 0.35 mm, as smaller diameters were deemed impractical to work with using our hand-made fabrication process.

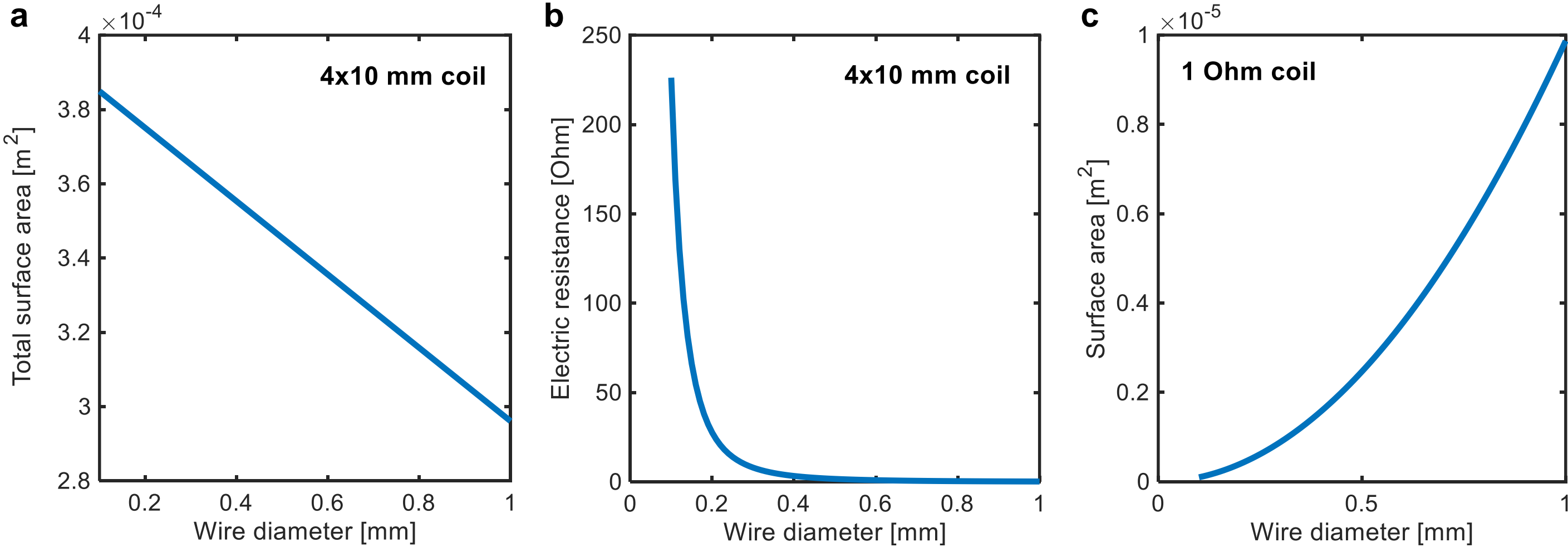

**Supplementary Fig. 4. Wire diameter evaluation**. **a** Coil surface area relative to the wire diameter considering constant external dimensions (4 mm for the external diameter and 10 mm length). **b** Electrical resistance relative to wire diameter considering constant external dimensions (4 mm external diameter and 10 mm length). **c** Surface area of a 1 Ω coil relative to different wire diameters.

### Coil wire length and electric requirements

In this study, the influence of different external coil diameters was not explored, so the coil diameter was fixed at 4 mm for all coil variants (differing only in length) to facilitate fabrication. Given an operating voltage $U$ = 12 V and current $I$ = 1.7 A, the coil's wire length $l_{wire}$ [m] is determined by:

$$l_{wire} = \frac{\pi U D^2}{4 I \rho_{el}} \tag{14}$$

Where $D = 3.5\times10^{-4}$ m is the wire diameter and $\rho_{el} = 1.45\times10^{-6}$ Ω.m is the wire resistivity. The resulting wire length, $l_{wire}$ = 0.468 m, corresponds to a coil with 37 turns assuming a 4 mm diameter. The maximum heat flux on the coil $q''_{coil}$ can be determined through:

$$q''_{coil} = \frac{UI}{\pi D l_{wire}} \tag{15}$$

Resulting in $q''_{coil} \approx 40$ kW/m$^2$, which is lower than the continuous limit of 64 kW/m$^2$ defined in Supplementary Note 4 as well as the CHF induced peak limit which is expected to occur at $q''_{coil} > 300$ kW/m$^2$. Finally, the coil length should be as long as possible to promote a more uniform heat distribution. In this case, the length was set at 40 mm (resulting in a pitch of 1.08 mm) to ensure compatibility with the shorter soft bending actuator variant. In summary, the coil is defined by the characteristics in Supplementary Table 9.

**Supplementary Table 9**. Coil characteristics.

| PARAMETER | ROOT DIAMETER | COIL LENGTH | PITCH | ELECTRIC RESISTANCE |
|---|---|---|---|---|
| VALUE | 4 mm | 40 mm | 1.08 mm | 7 Ω |

**Coil characteristics to maximize the rate of heat transfer**

If the maximization of the rate of heat transfer takes precedence over electrical requirements, then the coil can be optimized to maximize its surface area. The pressure chamber of the linear actuator is 150 mm long. Considering a maximum contraction of 25%, the heating element must span less than 112.5 mm, allowing 12.5 mm for the electrical connections. The maximum allowable coil length is 100 mm.

The coil pitch should be high enough to allow for reliable separation between each turn of wire, while small enough to guarantee the highest number of turns (and thus surface area) within the available volume. We selected a pitch equal to twice the wire diameter ($2 \times 0.35 = 0.70$ mm).

The number of wire turns in the coil is then $100/0.70 \approx 143$ turns, which at a diameter of 4 mm results in a wire length of $l_{wire} \approx 1.8$ m. The electrical resistance in this case is 27 Ω, which considering a maximum heat flux of 64 kW/m$^2$ allows for a peak power of 108 W, Supplementary Table 10. Note that, at these power settings, the actuator can achieve pressurization ratios over 100 kPa/s, which could lead to overpressure in one second or less. We advise against operating at these power settings if a control cycle frequency of at least 10 Hz cannot be guaranteed.

**Supplementary Table 10**. Coil characteristics.

| PARAMETER | ROOT DIAMETER | COIL LENGTH | PITCH | ELECTRIC RESISTANCE |
|---|---|---|---|---|
| VALUE | 4 mm | 100 mm | 0.7 mm | 27 Ω |

## Supplementary Note 3: FABRICATION OF SOFT STRUCTURES

### Bending actuator

The fabrication of the soft structure for the bending actuator follows a casting process[10] detailed step by step in Supplementary Table 11 and Supplementary Fig. 5.

**Supplementary Table 11**. Fabrication of the soft structure for the bending actuator.

| FABRICATION STEP | DESCRIPTION | FIGURE(S) |
|---|---|---|
| 1 | 3D-print the molds using a convenient rigid material such as PLA. 3D-print the soluble cores using water soluble PVA filament. CAD files available as supplementary materials:<br>1. sba_mold_left.stl<br>2. sba_mold_right.stl<br>3. sba_mold_extension.stl<br>4. sba_mold_tip.stl<br>5. sba_core.stl | Supplementary Fig. 5a |
| 2 | Clean the leftover residues from the printing process. Then, use four M3×35 screws and M3 nuts to assemble the mold with the core inside. | Supplementary Fig. 5b |
| 3 | Prepare 20 grams of Shore 40A platinum cured silicone (Smooth-Sil 940, Smooth-On, USA). Use a vacuum chamber to degas the uncured silicone for a period of ~2 min. Then, slowly pour the mixture into the mold, and let it cure at atmospheric conditions (20 to 25 ℃, 1 atm) for around 24 hours. | Supplementary Fig. 5c |
| 4 | Remove the screws from the mold. Then, align the mold extension and retighten the screws. Do not remove the actuator from the mold. | Supplementary Fig. 5d |
| 5 | Prepare 10 grams of platinum cured silicone (Eco-flex 00-30, Smooth-On, USA). Then, slowly pour the mixture into the mold extension and let it cure at atmospheric conditions. | Supplementary Fig. 5e |
| 6 | Remove the screws and the mold extension. Then, carefully separate the actuator from the mold. Avoid damaging the PVA core. | Supplementary Fig. 5f |
| 7 | Submerge the actuator in warm water (suggested 60℃) and wait for it to dissolve the sacrificial water-soluble core. To expedite the process, use a water pump to circulate water through the hollow PVA core. | Supplementary Fig. 5g |
| 8 | Prepare 6 grams of Shore 40A platinum cured silicone (Smooth-Sil 940, Smooth-On, USA). Slowly pour the mixture into the tip mold. Then, slowly fit the actuator into the mold and let it be cured at atmospheric conditions. | Supplementary Fig. 5h |
| 9 | Insert the heating element into the soft structure. Use cyanoacrylate glue to join the surfaces. Finally, insert a cable tie through the slots to further seal the joint. | Supplementary Fig. 5i |
| 10 | Using a syringe, fill the actuator with the desired working fluid through the feedback pressure tube. Then, connect the pressure sensor. At this point, the bending actuator is finished and ready to be used. | Supplementary Fig. 5j |

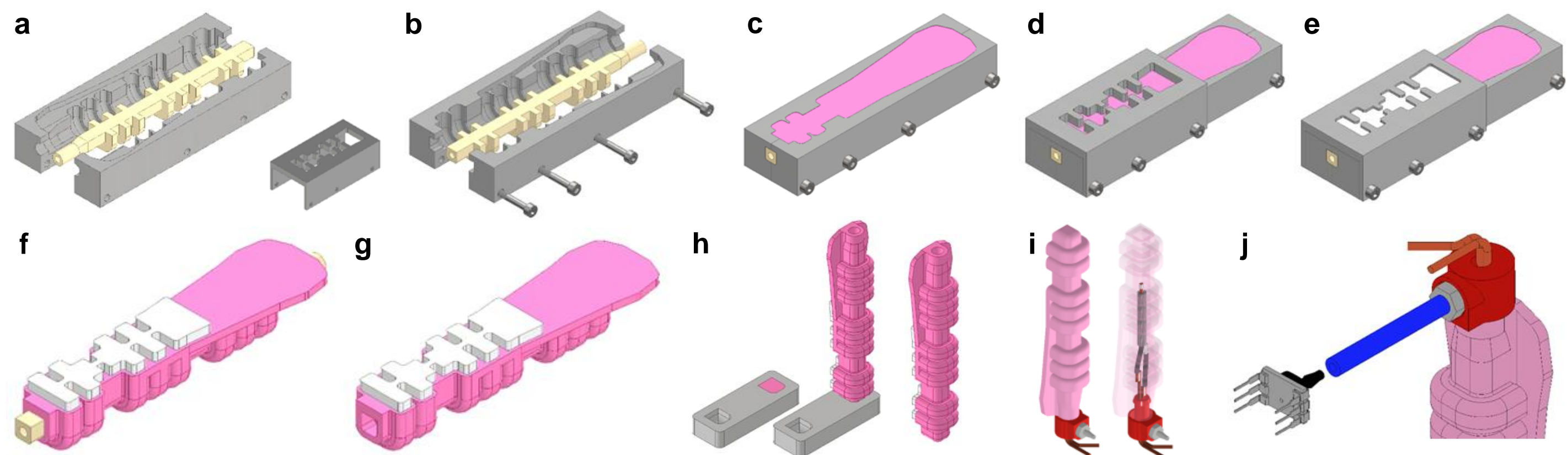

**Supplementary Fig. 5. Fabrication steps of the soft structure for the bending actuator**. **a** 3D-printing of molds and cores. **b** Assembly of the mold. **c** First silicone casting. **d** Assembly of the mold extension. **e** Second silicone casting to improve adhesion to surfaces. **f** Demolding. **g** Remove the PVA core. **h** Seal the actuator. **i** Assemble the heating element. **j** Fill the actuator with the working fluid (water in our actuators).

### Linear actuator

The fabrication of the soft structure for the linear actuator follows the process detailed step by step in Supplementary Table 12 and Supplementary Fig. 6.

**Supplementary Table 12**. Fabrication of the soft structure for the linear actuator.

| FABRICATION STEP | DESCRIPTION | FIGURE(S) |
|---|---|---|
| 1 | 3D-print the molds using a convenient rigid material such as PLA. 3D-print the terminal cap of the linear actuator using a thermally resistant material TPU 98A. Finally, 3D-print two anchor points for each actuator using a convenient rigid or soft material, we suggest PETG. CAD files available as supplementary materials:<br>1. linear_mold_top.stl<br>2. linear_mold_bottom.stl<br>3. linear_mold_center_cap.stl<br>4. linear_cap.stl<br>5. linear_anchor.stl | Supplementary Fig. 6a |
| 2 | Join the top and bottom halves of the mold. Use cable ties to secure/fix them together. | Supplementary Fig. 6b |
| 3 | Prepare 20 grams of platinum cured silicone (EcoFlex 00-30, Smooth-On, USA).<br>Use a vacuum chamber (VC2509AG, VacuumChambers, Poland) to degas the uncured silicone for a period of ~2 min. Then, slowly pour the mixture into the mold. Then, insert a 6 mm diameter metal extrusion/core into the mold, and center it with the plastic center cap. Let the mixture cure at atmospheric conditions (20 to 25 ºC, 1 atm) for 24 hours. | Supplementary Fig. 6c |
| 4 | Remove the center cap and the cable ties holding the mold together. Open the mold and remove the core with the silicone attached to it. Then, separate the silicone tube from the core and check the surface for defects. | Supplementary Fig. 6d |
| 5 | Cut the silicone tube to a length of 170 mm. Cut the 8 mm diameter braided sleeve to a length of 190 mm. | Supplementary Fig. 6e |
| 6 | Insert heating element into one end of the silicone tube. Use cyanoacrylate glue to join the surfaces. Then, insert the braided sleeve around the outside of the silicone tube, and secure it around the heating element using a cable tie.<br>Finally, trim the opposite end of the braided sleeve to the same length of the silicone tube. | Supplementary Fig. 6f |
| 7 | Insert the terminal cap into the open end of the silicone tube. Use cyanoacrylate glue to join the surfaces. Use a cable tie to further secure the joint. | Supplementary Fig. 6g |
| 8 | Use cyanoacrylate glue to join one anchor to each end terminal of the actuator. | Supplementary Fig. 6h |
| 9 | Using a syringe, fill the actuator with the desired working fluid through the feedback pressure tube. Then, connect the pressure sensor. At this point, the bending actuator is finished and ready to be used. | Supplementary Fig. 6i |

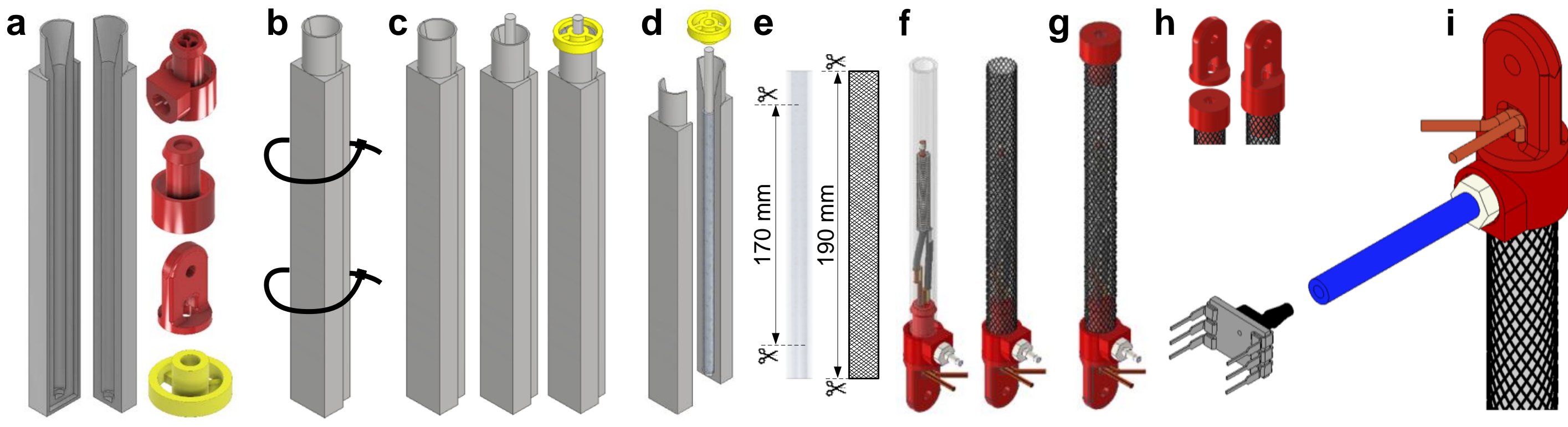

**Supplementary Fig. 6. Fabrication steps of the soft structure for the linear actuator**. **a** 3D-printing of molds and the components of the actuator. **b** Assemble the mold. **c** Silicone tube casting. **d** Demolding the silicone tube. **e** Trim the silicone tube and the braided sleeve. **f** Assemble the heating element. **g** Assemble the actuator components. **h** Assemble the actuator components. **i** Fill the actuator with the working fluid (water in our actuators).

**Supplementary Note 4: LIMIT OF HEAT FLUX TO ENSURE ACTUATOR'S SAFE OPERATION**

The effect of Critical Heat Flux (CHF) was discussed in the main text and represents the primary limiting factor for heat flux (rate of heat transfer per unit area) in liquid-gas phase transition actuators. However, to ensure the safe operation of the actuators, even in the event of leaks and structural failures, it is important to set a lower limit for sustained heat flux.

To assess this limit, a heating element was powered for 3 seconds and placed in direct contact with a silicone sample made of Ecoflex 00-30, Supplementary Fig. 7a. Four tests were conducted with electric currents of 1 A, 1.5 A, 2 A, and 2.5 A. Optical microscopy photographs shown in Supplementary Fig. 7 suggest an onset of thermal degradation starting at 36 kW/m², with more severe damage visible at 64 kW/m² and 100 kW/m². Light smoke was observed at 100 kW/m². Based on these observations, we selected 64 kW/m² as the upper limit for thermal surface load under normal operation. Supplementary Fig. 7g shows the silicone sample after contacting with the heating element coil with a current of 2.5 A (100 kW/m²). While the visible coil marks may not immediately compromise the actuator's functionality, they will degrade the soft structure over time. We conducted an experiment using a thermal camera to estimate the coil's surface temperature. Under natural convection, the surface temperature reached 190 ℃ after 3 s of operation at a heat flux of 64 kW/m², Supplementary Fig. 7h. Although the silicone manufacturer specifies a maximum useful temperature limit of 232 ℃, thermal degradation was observed at 64 kW/m². This indicates that, despite the contact area between the coil and the silicone being less than 0.5 mm² per turn, it is sufficient to create a hotspot on the coil with an overtemperature of at least 40 ℃ compared to the surrounding coil area. These findings suggest that this effect is more significant than previously anticipated and that heat flux limits could potentially be increased by optimizing the actuator design to eliminate direct contact between the coil and the silicone.

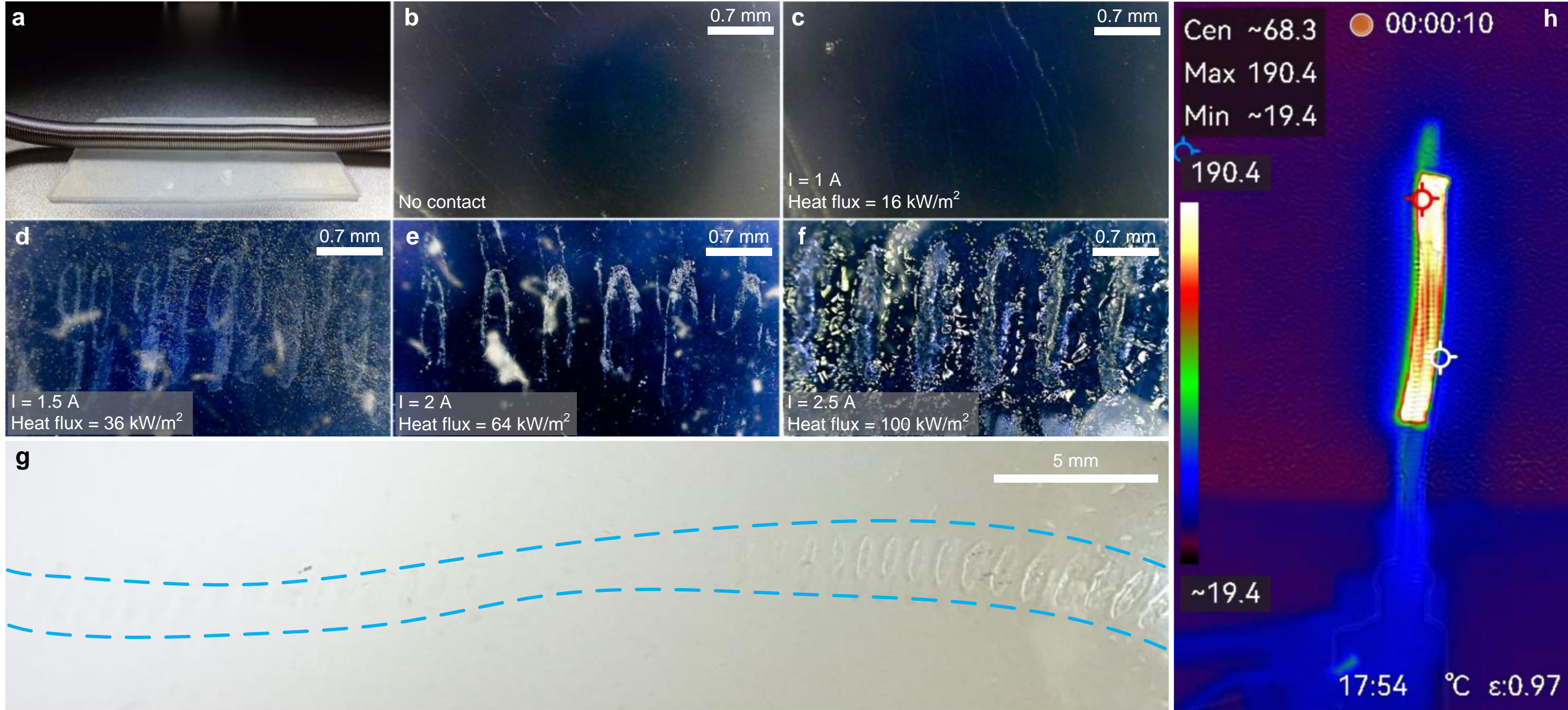

**Supplementary Fig. 7. Contact of the heating element with the silicone soft structure**. **a** Heating element coil in direct contact with a silicone sample made of Ecoflex 00-30. **b** Optical microscopy photograph of silicone sample without coil contact. **c** Optical microscopy photograph of silicone sample with coil contact with current of 1 A. **d** Optical microscopy photograph of silicone sample with coil contact with current of 1.5 A. **e** Optical microscopy photograph of silicone sample with coil contact with current of 2 A. **f** Optical microscopy photograph of silicone sample with coil contact with current of 2.5 A. **g** Silicone sample after contacting with the heating element coil with a current of 2.5 A. **h** Coil's surface temperature captured by a thermal camera (Pocket2, HIKMICRO, China).

**Supplementary Note 5: BRAIDED SLEEVE**

Most commercially available braided sleeves follow one of the four braiding patterns shown in Supplementary Fig. 8. Triaxial braided sleeves are generally not suitable for use in linear actuators, as some of their inextensible yarns run parallel to the sleeve's longitudinal axis, Supplementary Fig. 8a. The biaxial braided sleeves are all suitable alternatives, Supplementary Figs. 8b to 8d. The actuator's maximum displacement is proportional to the sleeve's own extensibility, i.e., the maximum longitudinal displacement, so it is useful to understand how different braid parameters affect extensibility. By approximating the path of each yarn to a cylindrical helix, i.e., ignoring the weaving curvature around each crosspoint, the total length $L$ of the braided sleeve can be determined by:

$$L = b \times cos(\theta) \tag{11}$$

Where $b$ is the length of each yarn and $\theta$ is the braid angle formed between each yarn and the sleeve's longitudinal axis. Since the yarn length is a fixed parameter, the sleeve's longitudinal extensibility may be approximated by determining the upper and lower limits of the variable braid angle. The longitudinal force done by the braid[69], assuming the braid remains cylindrical and without considering internal friction, when subjected to an internal pressure can be approximated by:

$$F = \frac{Pb^2}{4\pi n^2}\left(3 \times cos^2(\theta) - 1\right) \tag{12}$$

Where $P$ is the relative pressure acting on the inner side of the braid and $n$ is the number of turns each yarn makes around the braid's longitudinal axis. Solving Supplementary Equation (12) for $\theta$ when $F$=0 results in a first positive root of 54.7º, known as the "*neutral angle*". This represents a theoretical upper limit for our braid angle during actuator contraction and is independent of other braid parameters. A lower limit can also be approximated by analyzing the fraction of area covered by the braid, known as Cover Factor $C_F$, which can be calculated through:

$$C_F = 1 - \left(1 - \frac{n \times W_y \times N_y}{b \times sin(\theta) \times cos(\theta)}\right)^2 \quad (13)$$

Where $W_y$ is the width of each yarn and $N_y$ is the number of yarns forming the braid. Analyzing a unitary braid cell such as the one schematized in Supplementary Fig. 8e, the cover factor would be $C_F$ = 1-*(EFGH/ABCD)*, with *EFGH* and *ABCD* being the areas of the corresponding polygons. When a braid is extended, the exposed area of each cell (*EFGH*) decreases as the braid angle becomes smaller until approximately no area is left exposed, $C_F \approx 1$. At this point, adjacent yarns contact each other, preventing the braid's diameter from further reducing and, consequently, impeding the braid's length from increasing. Through Supplementary Equation (13) we verify that braids with lower number of yarns $N_y$ and/or lower yarn widths $W_y$ have lower $C_F$ values for a given braid angle, resulting in more contractible braids. One last consideration is that the higher the maximum exposed area within each cell, which occurs at $\theta$ = 45º, the greater the risk of the actuator's elastomer soft structure bubbling through the braid and rupturing.

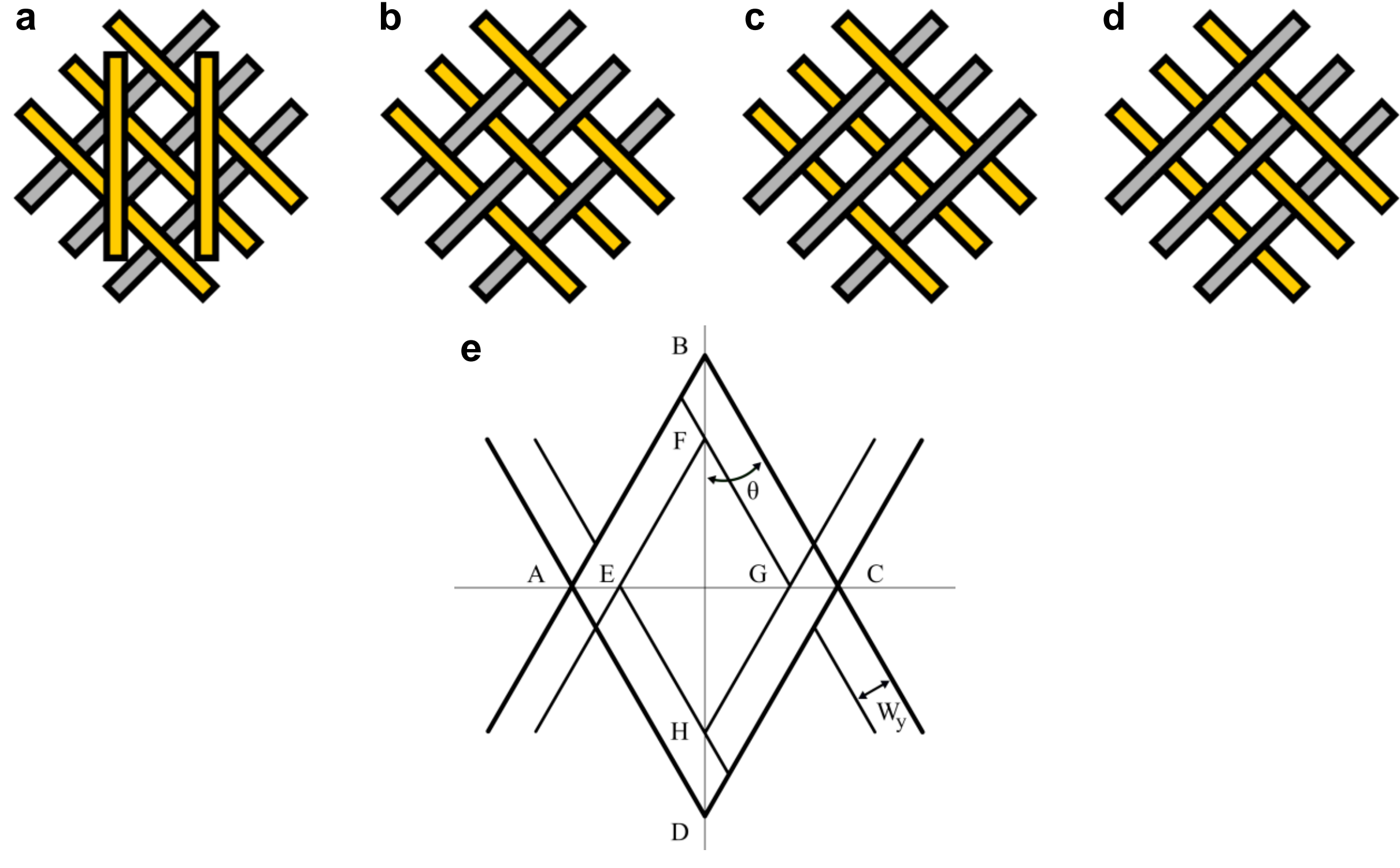

**Supplementary Fig. 8. Braided sleeve patterns and unitary braid cell**. **a** Triaxial braided sleeve pattern. **b** Biaxial braided sleeve pattern. **c** Biaxial braided sleeve pattern. **d** Biaxial braided sleeve pattern. **e** Unitary braid cell.

## Supplementary Note 6: ACTUATOR STATIC MODEL

### Ideal McKibben actuator

Considering an ideal McKibben pneumatic artificial muscle, characterized by:

1. Zero static and dynamic friction;
2. Inflatable bladder made up of an infinitely soft material (with zero Young's modulus);
3. Inflatable bladder made up of an infinitely thin material (with zero thickness);
4. Braided sleeve made up of infinitely inextensible strands.

The output force of such actuator, as described by Chou and Hannaford[68], can be calculated through:

$$F = \frac{Pb^2}{4\pi n^2}\left(3 \times \cos^2(\theta) - 1\right) \tag{15}$$

Where $P$ [Pa] is the relative internal pressure, $b$ [m] is the fixed length of the braid's yarns, $n$ is the number of turns each yarn makes around the braid's longitudinal axis, and $\theta$ is the variable braid angle. The braid angle is a function of the braid's current length, as described by Supplementary Equation (11). From Supplementary Equation (15) and Supplementary Equation (11) we have:

$$P_{ideal} = F\,\frac{4\pi n^2}{3L^2 - b^2} \tag{16}$$

### Phase-change actuator isotonic data

The pressure-displacement relation of our actuators under zero load (free stroke) isotonic tests can be closely approximated by a 2nd degree polynomial. Supplementary Fig. 9 shows data recorded during free stroke test of an EcoFlex 00-30 actuator, as well as a conveniently fitted polynomial:

$$P(\Delta L) = A(\Delta L)^2 + B(\Delta L) + C \tag{17}$$

Where the factors A, B and C are in Supplementary Table 13, and $\Delta L$ [m] is the linear displacement of the actuator.

**Supplementary Table 13**. Factors for 2nd degree polynomial fitting of isotonic test.

| FACTOR | A | B | C |
|---|---|---|---|
| VALUE | 1.563e+07 | 7.608e+05 | 3677 |

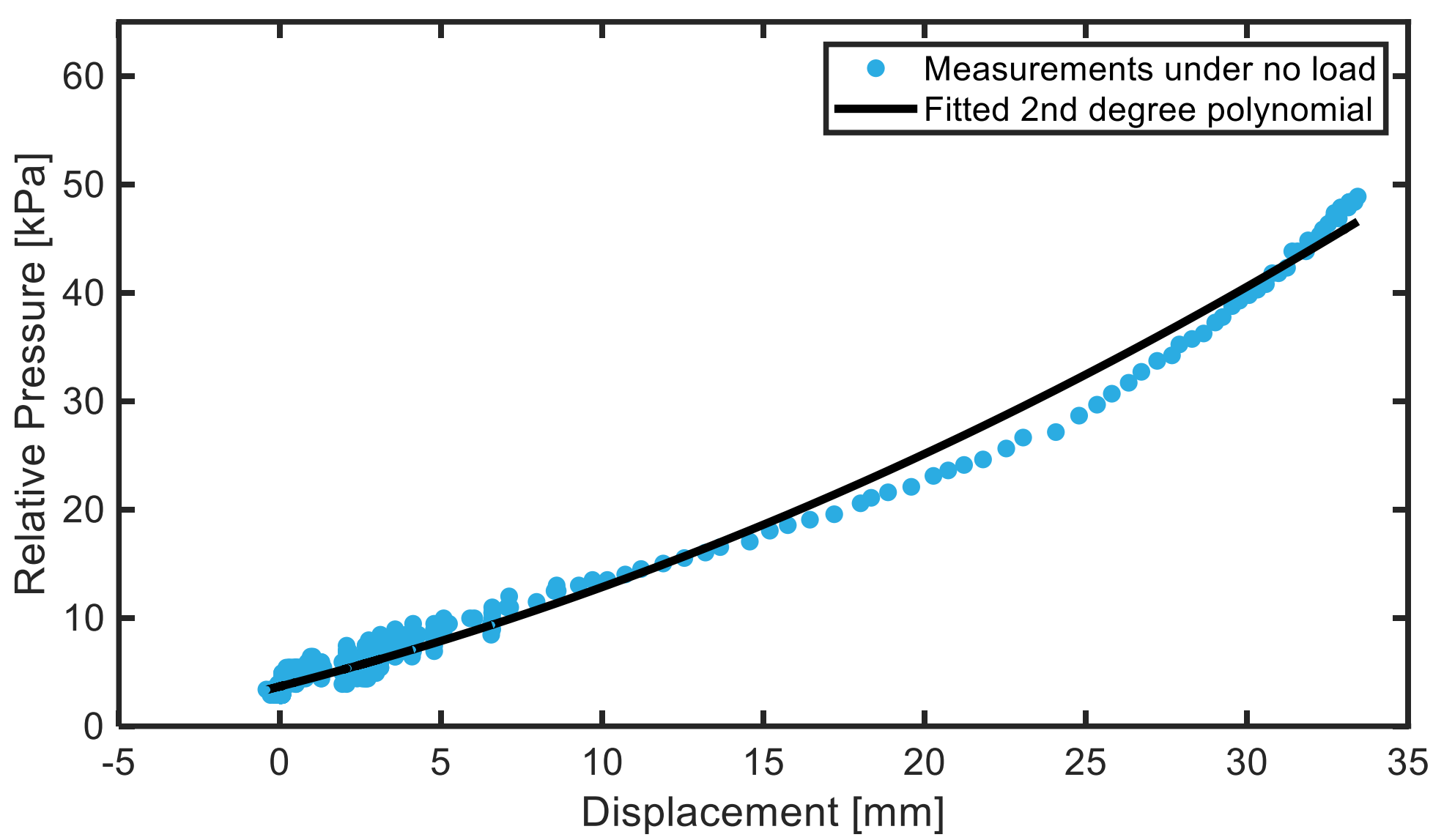

**Supplementary Fig. 9. Pressure-displacement of free stroke test fitted with a 2nd degree polynomial curve**. **The actuator is made of Ecoflex 00-30.**

**Semi-empirical static model**

The pressure necessary to deform the actuator is accounted by the empirical model in Supplementary Equation (17). The overpressure necessary to output a given force is accounted for by the ideal static model in Supplementary Equation (16). Merging the two models together results:

$$P\left(F,\Delta L\right)=F\frac{4\pi n^{2}}{3L^{2}-b^{2}}+A(\Delta L)^{2}+B\Delta L+C \tag{18}$$

Note that Coulomb friction was not considered in Supplementary Equation (18). An extra term accounting for this may be necessary in applications demanding relatively high load and/or high precision.

### Validation of the model

To validate the model, we use a different data set recorded during an isotonic test of the same actuator, operating under a constant load of 10.8 N, Supplementary Fig. 10. The model was able to approximate the internal pressure required to achieve a set displacement value within ± 5 kPa.

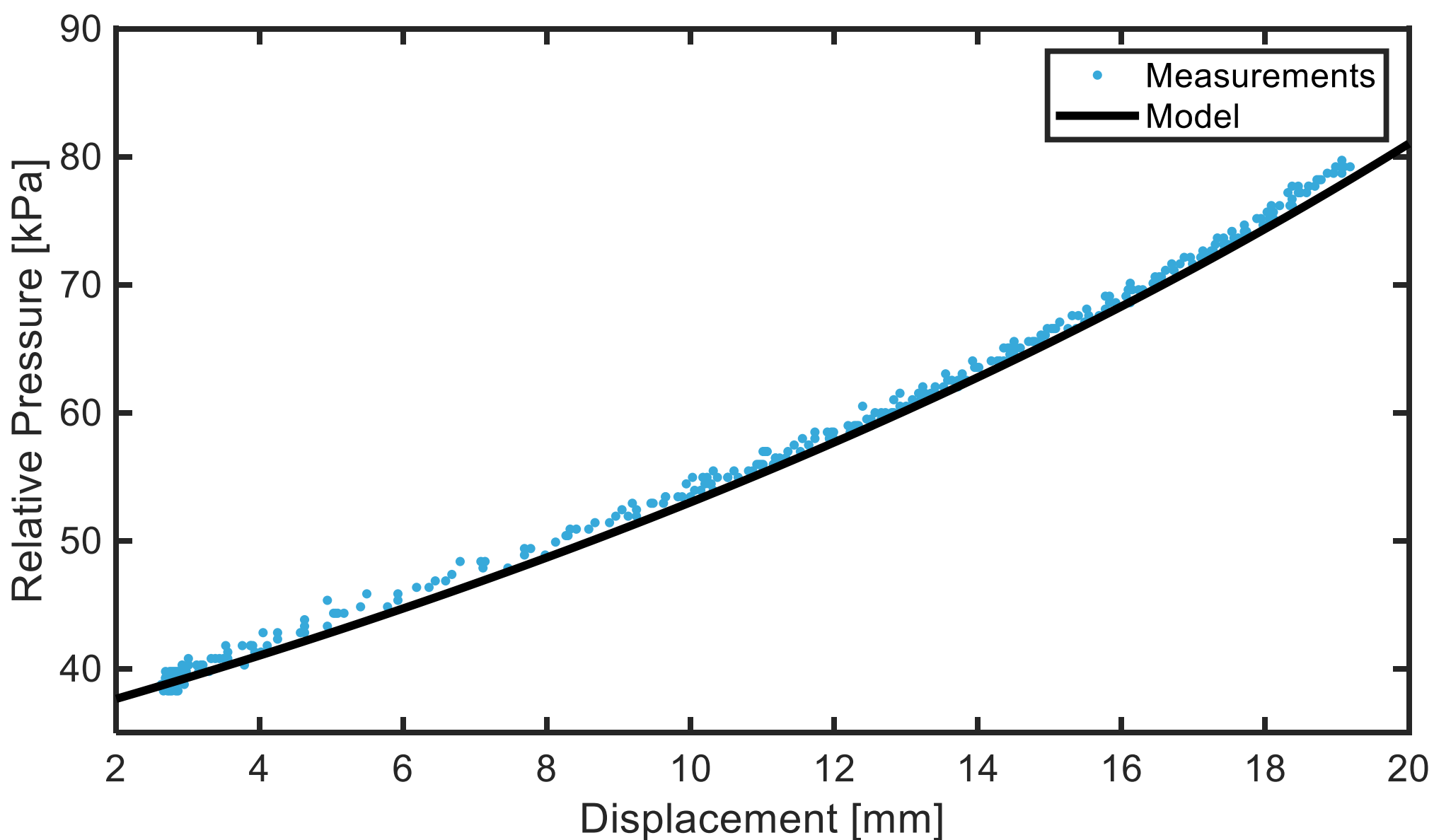

**Supplementary Fig. 10. Pressure-displacement of 10.8 N isotonic test fitted with a 2nd degree polynomial curve, presented for both the model and experimental data. The actuator is made of Ecoflex 00-30.**

## Supplementary Note 7: ENERGY ANALYSIS

To obtain energy performance data, including work output, power output and energy efficiency, a series of isotonic tests were conducted. Actuators were preloaded with loads ranging from 4 N to 20 N, in increments of 2 N, and then pressurized up to 130 kPa at a constant power level. To achieve a more realistic assessment of actual work output capability, unconstrained by our empirically defined safety limit of 130 kPa, additional tests were conducted at pressures up to 200 kPa. Furthermore, to determine the effect of input power on efficiency, all tests were repeated at two different power levels, 25 W and 50 W, Supplementary Figs. 11a to 11c.

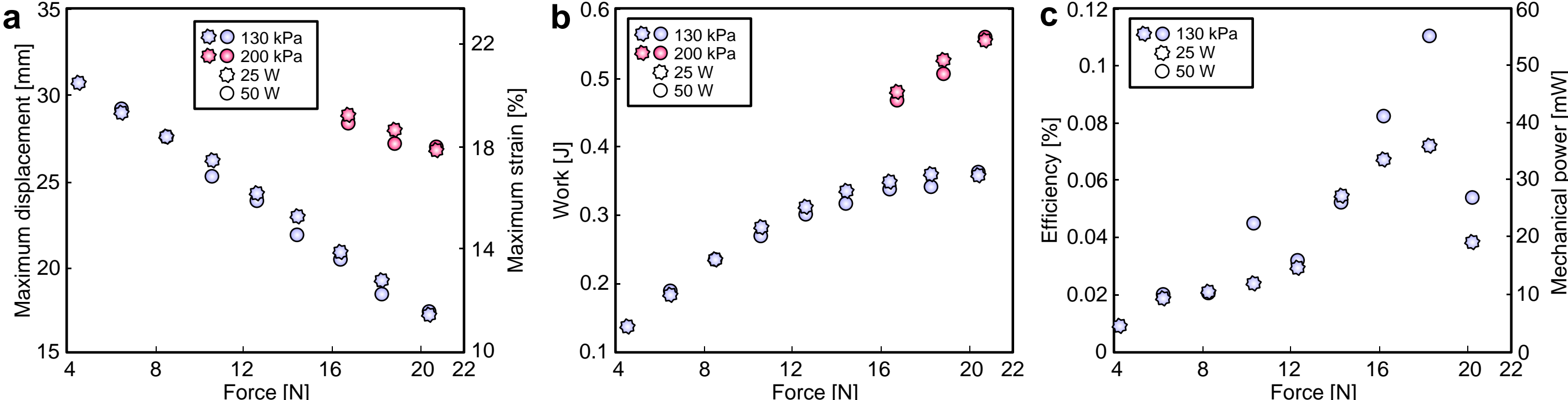

**Supplementary Fig. 11. Isotonic tests to obtain energy performance data: work output, power output and energy efficiency**. **a** Maximum linear displacement reached at 130 kPa and 200 kPa under different isotonic loads. **b** Work output after pressurization to 130 kPa and 200 kPa under different isotonic loads. **c** Actuator efficiency and mechanical power.

The mechanical work output was then calculated based on isotonic data by multiplying each test's maximum displacement with their respective pre-load, Supplementary Fig. 11b. The specific work obtained in these tests is approximately 30 J/kg, situating phase transition technology within the range of specific work values expected from high-voltage electrostatic actuators, including HASEL actuators and DEAs. Actuation efficiency was then calculated for each test:

$$\eta = \frac{W_{out}}{E_{in}} = \frac{W_{out}}{V \times I \times \Delta t} \tag{19}$$

Where $\eta$ is the energy efficiency, $W_{out}$ [J] is the mechanical work output during actuator contraction, $E_{in}$ [J] is the electric energy input during actuator contraction, $V$ [V] is the voltage, $I$ [A] is the electric current, and $\Delta t$ [s] is the time period required for actuator contraction.

A maximum efficiency value of 0.11% was reached under a load of 18 N at 50 W, Supplementary Fig. 11c, suggesting that operation at increased power settings results in increased efficiency. This value aligns with previously reported values for liquid-gas phase transition actuators[53,56]. The increase in efficiency at higher power settings is attributed to faster actuation cycles, which reduce the time available for energy dissipation, thereby enhancing efficiency. Another important conclusion is that power settings do not affect the force value at which actuators reach peak efficiency. This can be observed in Supplementary Fig. 11c, where peak efficiency was achieved at 18 N at both 25 W and 50 W. In Supplementary Table 14, we aggregate performance figures reported for various liquid-gas phase transition actuators. In Supplementary Table 15, we extend the search to include other soft actuation technologies.

**Supplementary Table 14**. Performance of various liquid-gas phase transition actuators.

| ACTUATOR TECHNOLOGY | SPECIFIC ENERGY [J/kg] | STRAIN RATE [%/s] | MAX. STRAIN [%] | FORCE [N] | OPERATING VOLTAGE [V] | INPUT POWER [W] | EFFICIENCY [%] |
|---|---|---|---|---|---|---|---|
| **Ours**: Liquid-gas phase transition | **30** | **16.6** | **27** | **52** | **24** | **10-100** | **0.11** |
| Liquid-gas phase transition[55] | – | 1.48 * (estimated) | 36 | 50 | 17.35 | 15 | – |
| Liquid-gas phase transition[56] | – | – | – | 60 | <30 | <45 | 0.2[b] |
| Liquid-gas phase transition[54] | 10.55 * (estimated) | 0.46 * (under load) | 10 * (under load) | 300 | 15 | ~29 | 0.032[c] |
| Liquid-gas phase transition[53] | 40[a] | 1.2 | 20 | >31 | >8 | up to 900 | <<1[b] |

(a) Estimated from reported energy density data under the assumption of actuator density ~1000 kg/m$^3$.
(b) Calculated assuming the time period required by the actuator to reach maximum contraction under a constant load is equal to the time period required by the actuator to reach a blocked force equal to the constant load.
(c) Estimated from data reported in the paper regarding the lifting of a 4.3kg load.
(d) Not further specified by the author.

**Supplementary Table 15**. Performance of electrothermal, electrochemical and electrostatic actuators.

| ACTUATOR TECHNOLOGY | SPECIFIC ENERGY [J/kg] | STRAIN RATE [%/s] | MAX. STRAIN [%] | FORCE [N] | OPERATING VOLTAGE [V] | INPUT POWER [W] | EFFICIENCY [%] |
|---|---|---|---|---|---|---|---|
| Electrothermal TCA[49] | 241 | 13.75 | 55 | 0.78 | 28 | up to 10 | — |
| Electrothermal TCA (Coiled Polymer Fiber)[76] | 2480 | 18 - 100 * (18 under 7.2 kg load) | 49 | > 70 | 30 per cm | | 1.32 |
| Electrothermal LCE[50] | — | 1-2 * (estimated) | 41 | 3.92 | 3 | — | — |
| Electrochemical SECA[41] | — | — | 150 | 9.74 | <3 | 1 -2 | <0.8 |
| Electrochemical Liquid Metal Actuator[47] | 0.336[a] | — | 31.5 | <0.005 | 1-3 | — | — |
| Electrostatic DEA[36] | 12.9 | — | 30[b] | 32 | 4200 | — | — |
| Electrostatic DEA[77] | 19.8 | — | 24 | 5 | 3750 | — | 1.5 |
| Electrostatic EBM[38] | 1.16 | 1250 | 42.8 | <6 | < 8000 | — | 20[c] |
| Electrostatic HASEL[37] | 70 | 145 * (estimated) | 79[d] | 14.715 | 20000 to 30000 | — | 21 |

[a] Estimated using reported energy density (2100 J/m$^3$) and liquid metal density (6250 kg/m$^3$).
[b] Higher value of 46% reported for actuators without fixing hardware.
[c] When working as a volumetric pump, not as an actuator.
[d] Values up to 124% were reported when actuating near resonant frequency.

**Supplementary Note 8: SOFT QUADRUPED ROBOT BIXO**

The Bixo robot prototype is composed of multiple mechatronic components, Supplementary Table 16 and Supplementary Fig. 12. To facilitate the reproducibility of the robot, the CAD files of structural components are made available. These components can be 3D printed in a desktop FFF machine. Supplementary Table 17 and Supplementary Fig. 13 detail each assembly step of the robot. The MATLAB Graphical User Interface (GUI) used to monitor and control the robot is in Supplementary Fig. 14, while Supplementary Fig. 15 shows the robot's wiring diagram.

**Supplementary Table 16**. Bixo robot components.

| COMPONENT NUMBER AND DESCRIPTION | CAD FILE | MATERIAL | QUANTITY | FIGURE(S) |
|---|---|---|---|---|
| BX1: Spar | spar.stl | ABS (suggested) | 2 | Supplementary Fig. 12a |
| BX2: Fork | fork.stl | ABS (suggested) | 2 | Supplementary Fig. 12b |
| BX3: Right socket | right_socket.stl | ABS (suggested) | 2 | Supplementary Fig. 12c |
| BX4: Left socket | left_socket.stl | ABS (suggested) | 2 | Supplementary Fig. 12d |
| BX5: Top bridge | top_bridge.stl | ABS (suggested) | 1 | Supplementary Fig. 12e |
| BX6: Lead screw nut | linear_nut.stl | ABS (suggested) | 2 | Supplementary Fig. 12f |
| BX7: Center bridge | mid_bridge.stl | ABS (suggested) | 1 | Supplementary Fig. 12g |
| BX8: Lower bridge | low_bridge.stl | ABS (suggested) | 1 | Supplementary Fig. 12h |
| BX9: Lead screw | linear_screw.stl | ABS (suggested) | 1 | Supplementary Fig. 12i |
| BX10: Bending actuator | - | - | 4 | Supplementary Fig. 12j |
| BX11: 6×6 mm Switch | - | - | 2 | Supplementary Fig. 12k |
| BX12: M2×16 Screw | - | - | 5 | Supplementary Fig. 12l |
| BX13: 250:1 Micro Metal Gearmotor | - | - | 1 | Supplementary Fig. 12m |
| BX14: M1.6×4 Screw | - | - | 2 | Supplementary Fig. 12n |
| BX15: M2×6 Screw | - | - | 5 | Supplementary Fig. 12o |

| BX16: M2 Hex nut | - | - | 4 | Supplementary Fig. 12p |
|---|---|---|---|---|

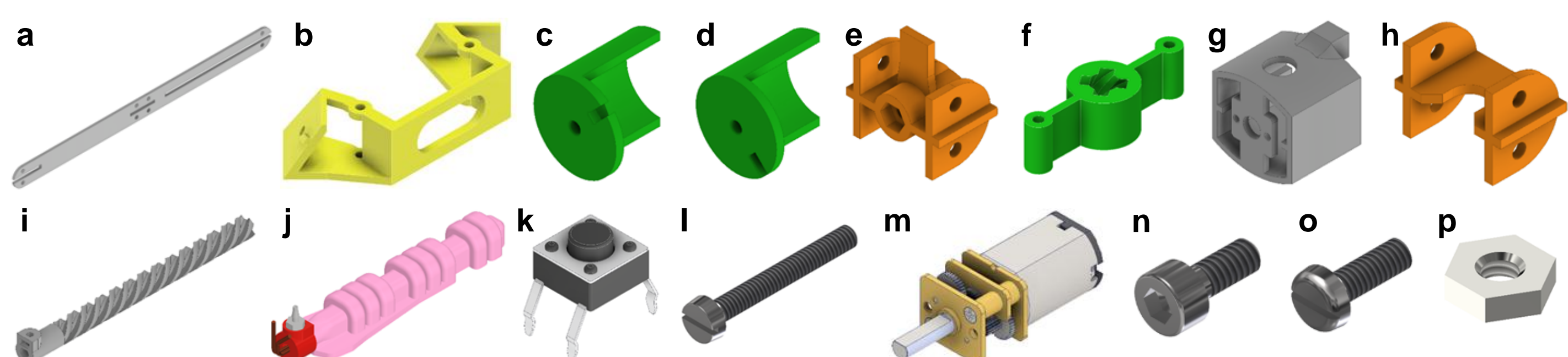

**Supplementary Fig. 12. Components to assemble the soft quadruped robot Bixo**. **a** Spar. **b** Fork. **c** Right socket. **d** Left socket. **e** Top bridge. **f** Lead screw nut. **g** Center bridge. **h** Lower bridge. **i** Lead screw. **j** Bending actuator. **k** Switch. **l** M2×16 screw. **m** 250:1 micro metal gearmotor LP 6V. **n** M1.6×4 screw. **o** M2×6 screw. **p** M2 hex nut.

**Supplementary Table 17**. Assembly steps for the robot Bixo.

| ASSEMBLY STEP | DESCRIPTION | FIGURE(S) |
|---|---|---|
| 1 | Assemble the geared motor BX13 to the center bridge BX7 using two M1.6×4 screws BX14. | Supplementary Fig. 13a |
| 2 | Align the central slots of both spars BX1 to the interior grooves of the center bridge BX7. Then, glue them together using cyanoacrylate glue. | Supplementary Fig. 13b |
| 3 | Insert one lead screw nut BX6 to approximately the center of the lead screw BX9. | Supplementary Fig. 13c |
| 4 | Connect the lead screw to the axis of the geared motor using one M2×6 screw BX15 and one M2 hex nut BX16. Use the hole on the central bridge to access and tighten the screw. | Supplementary Fig. 13d |
| 5 | Insert one M2×16 screw BX12 through the center hole of the top bridge BX5. Secure the screw using one M2 hex nut BX16. | Supplementary Fig. 13e |
| 6 | Align the M2×16 screw on the top bridge with the hole in the linear screw. Then, secure the top bridge to the spars using four M2×6 screws BX15 and M2 hex nuts BX16. | Supplementary Fig. 13f |
| 7 | Insert one linear nut BX6 through the lower slots of the spars. Then, insert the lower bridge BX8 through the same slots and secure it in place using four M2×6 screws BX15 and M2 hex nuts BX16. | Supplementary Fig. 13g |
| 8 | Insert one M2 hex nut in each right socket BX3 and left socket BX4. | Supplementary Fig. 13h |
| 9 | Secure the soft actuators BX10 to each socket using cyanoacrylate glue. Allow for a 2 mm gap between the actuator and the hex nut on the socket. | Supplementary Fig. 13i |
| 10 | Assemble the sockets (with the actuators attached) on the forks BX2 using M2×6 screws. | Supplementary Fig. 13j |
| 11 | Assemble the forks (with the actuators attached) on the lead screw nuts BX6 using four M2×16 screws BX12 and M2 nuts BX16. | Supplementary Fig. 13k |
| 12 | Attach the limit switches BX11 on both the center and top bridges using cyanoacrylate glue. At this point, the robot is ready to be connected to the controller. Please refer to the | Supplementary Fig. 13l |
| | suggested wiring diagram in Supplementary Fig. 14. | |

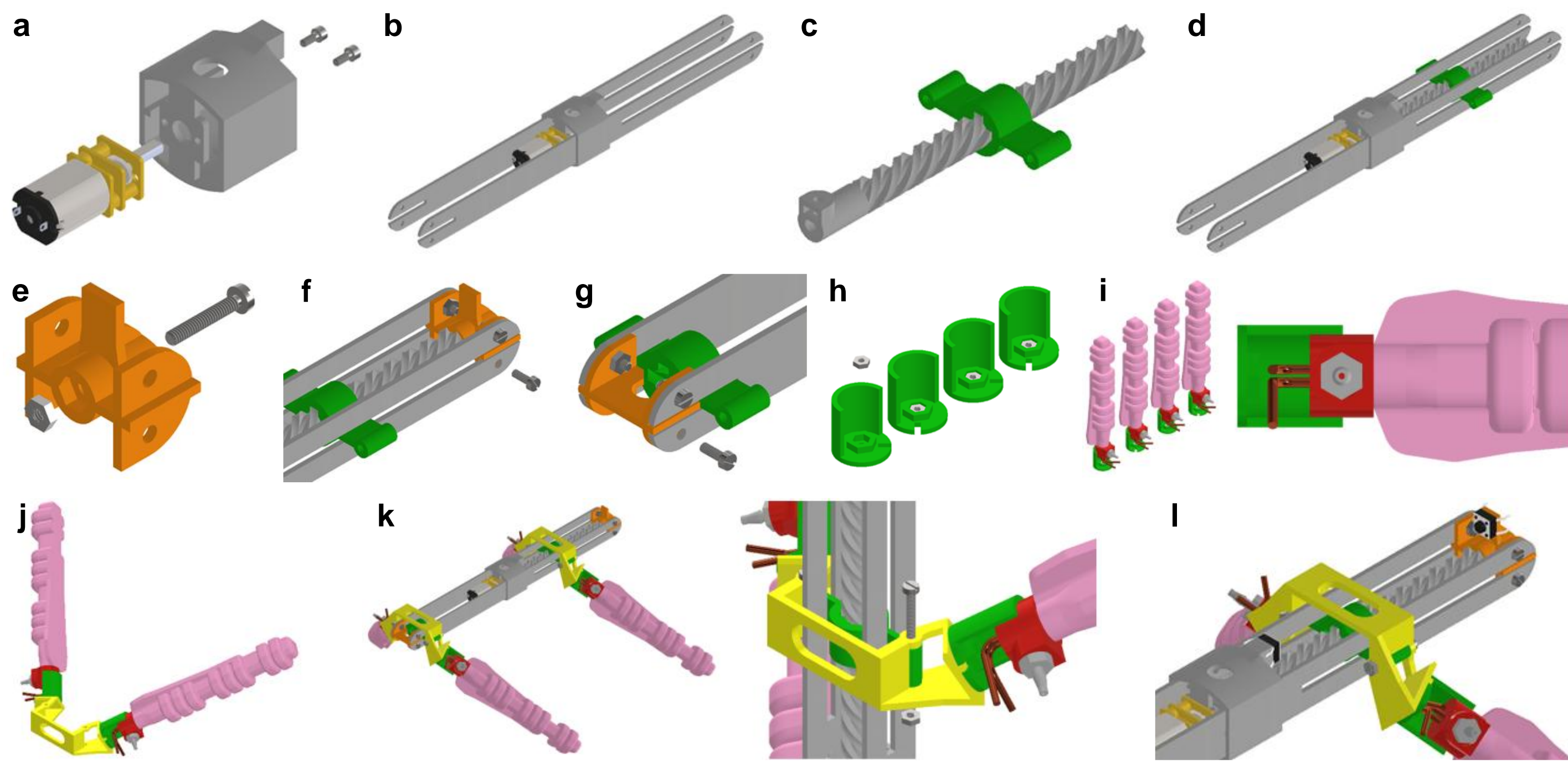

**Supplementary Fig. 13. Assembly steps of the soft quadruped robot Bixo**. **a** Assembly step 1. **b** Assembly step 2. **c** Assembly step 3. **d** Assembly step 4. **e** Assembly step 5. **f** Assembly step 6. **g** Assembly step 7. **h** Assembly step 8. **i** Assembly step 9. **j** Assembly step 10. **k** Assembly step 11. **l** Assembly step 12.

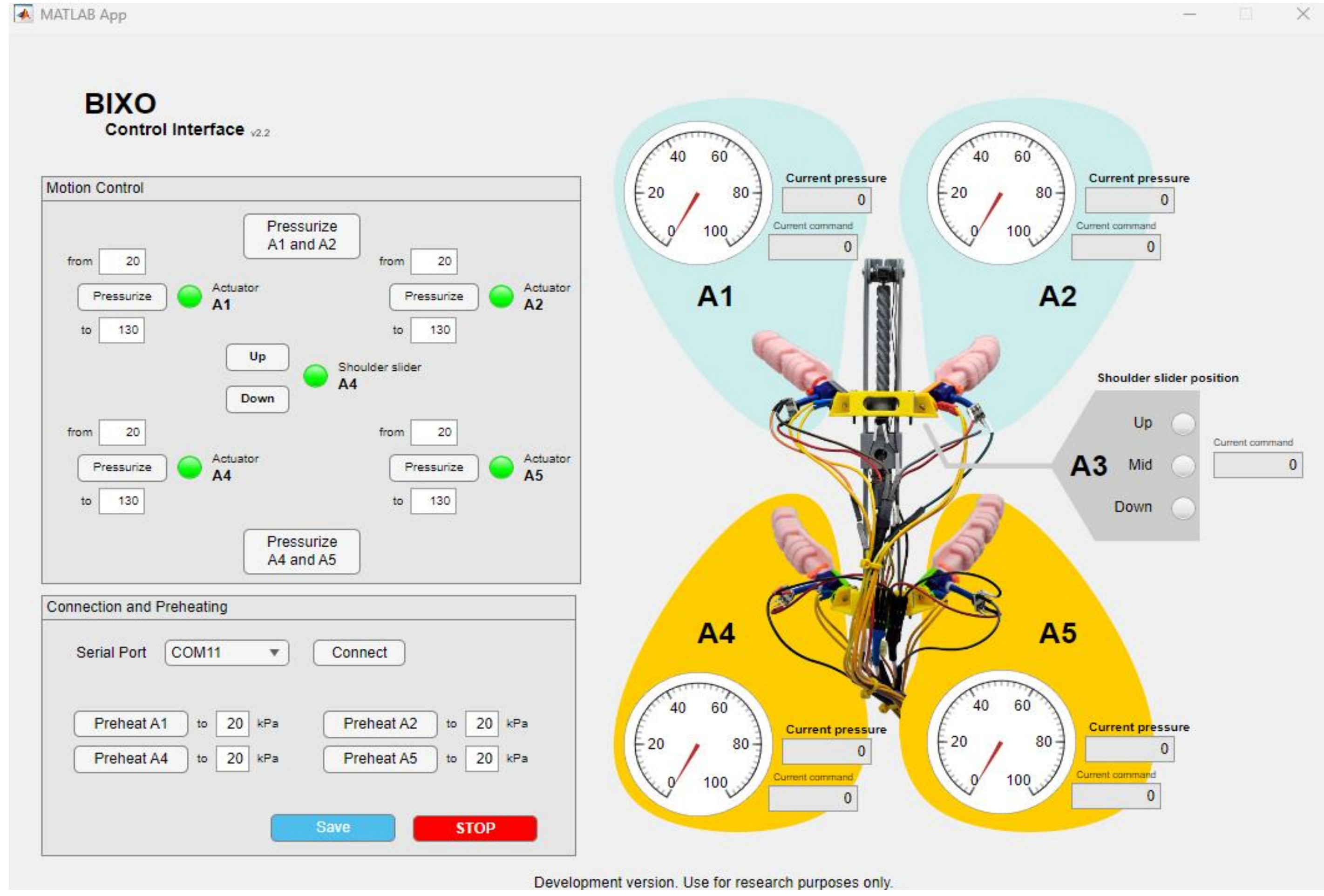

**Supplementary Fig. 14. Graphical User Interface (GUI) used to monitor and control the Bixo robot**.

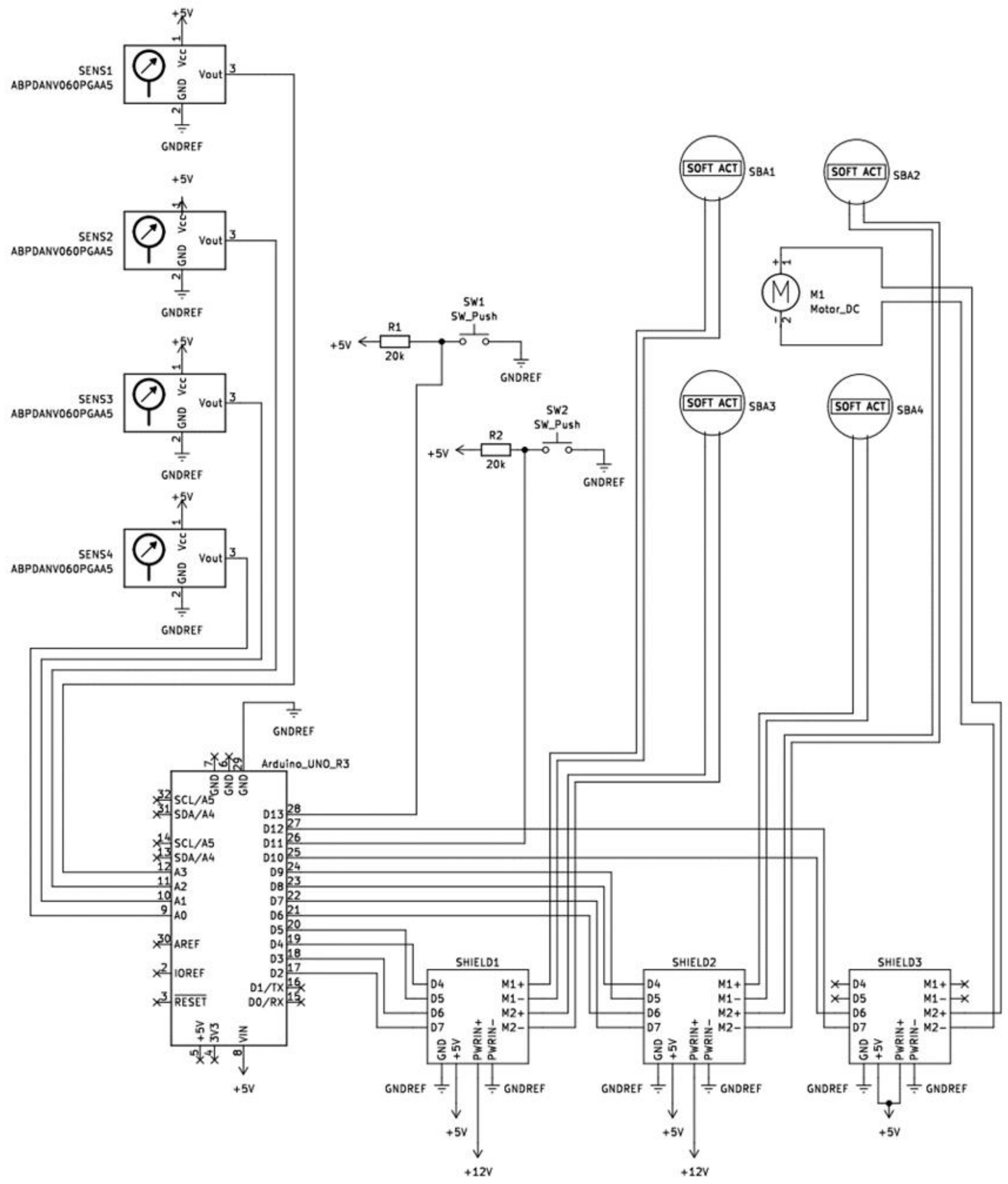

**Supplementary Fig. 15. Wiring diagram of the soft quadruped robot Bixo**.